\documentclass[journal]{new-aiaa}

\usepackage{mathtools,amsmath}
\usepackage{epstopdf}
\usepackage{gensymb}
\usepackage{bbold}
\usepackage{array,multirow,rotating}
\usepackage{ifthen}
\usepackage{varioref}
\usepackage{wrapfig}
\usepackage{threeparttable}
\usepackage{dcolumn}
\newcolumntype{d}{D{.}{.}{-1}}
\usepackage{nomencl}
\usepackage{makeidx}
\makenomenclature
\usepackage{fancyvrb}
\usepackage{lettrine}
\usepackage{hyperref}
\usepackage{footnote}
\usepackage{fancyhdr}
\usepackage{empheq}
\usepackage{pifont}
\usepackage{subcaption}

\usepackage[super]{nth}

\usepackage[dvipsnames]{xcolor}
\usepackage{color, colortbl}
\usepackage{mwe}
\usepackage{algorithm}

\hypersetup{
  colorlinks   = , 
  urlcolor     = black, 
  linkcolor    = blue, 
  citecolor   = red 
}

\usepackage{url}
\usepackage{breakurl}
\usepackage{longtable,tabularx}
\title{Improved Aircraft Environmental Impact Segmentation via Metric Learning}

\author{
 	Zhenyu Gao\footnote{Postdoctoral Fellow, Department of Aerospace Engineering and Engineering Mechanics, The University of Texas at Austin, AIAA Member}\\
        {\itshape Department of Aerospace Engineering and Engineering Mechanics\\The University of Texas at Austin, Austin, TX 78712}\\
        \vspace{0.5cm}
        \upshape Dimitri N. Mavris\footnote{S.P. Langley Distinguished Regents Professor, School of Aerospace Engineering, Georgia Institute of Technology, AIAA Fellow}\\
        {\itshape School of Aerospace Engineering, Georgia Institute of Technology, Atlanta, GA 30332}\\
  }

\begin{document}


\maketitle

\begin{abstract}

Accurate modeling of aircraft environmental impact is pivotal to the design of operational procedures and policies to mitigate negative aviation environmental impact. Aircraft environmental impact segmentation is a process which clusters aircraft types that have similar environmental impact characteristics based on a set of aircraft features. This practice helps model a large population of aircraft types with insufficient aircraft noise and performance models and contributes to better understanding of aviation environmental impact. Through measuring the similarity between aircraft types, distance metric is the kernel of aircraft segmentation. Traditional ways of aircraft segmentation use plain distance metrics and assign equal weight to all features in an unsupervised clustering process. In this work, we utilize weakly-supervised metric learning and partial information on aircraft fuel burn, emissions, and noise to learn weighted distance metrics for aircraft environmental impact segmentation. We show in a comprehensive case study that the tailored distance metrics can indeed make aircraft segmentation better reflect the actual environmental impact of aircraft. The metric learning approach can help refine a number of similar data-driven analytical studies in aviation.

\end{abstract}

\section{Introduction}\label{sec:intro}

The global air transportation system has grown to become an integral and indispensable part of the global economy. With the projected growth in scale over the coming decades, the aviation industry is expected to better facilitate the movement of people and cargo around the globe in more diversified forms. In the meantime, however, the environmental impact of aviation, also referred to by some as the most significant adverse impact of aviation~\citep{waitz2004aviation}, has materialized as an enormous international concern. The three primary facets of negative aviation environmental impacts are: (1) local air quality impacts, (2) climate change impacts, and (3) community noise impacts~\citep{faa2015aviation}. If not properly addressed, these environmental impacts can exacerbate health-harming air pollution,  accelerate global warming, and undermine affected population's mental well-being. It is a consensus that the aviation industry must keep booming to meet future economic needs while simultaneously operating harmoniously within the constraints imposed by clean air and water, limited noise impacts, and a livable climate. Therefore, sustainable aviation is among the most crucial research themes in aerospace and transportation engineering. In a broad sense, sustainable aviation has two branches. The first branch focuses on the development of advanced low emissions and noise technologies, such as More Electric Aircraft (MEA)~\citep{rosero2007moving}, advanced engine technologies~\citep{hughes2011aircraft}, and alternative fuels~\citep{blakey2011aviation}. The second branch focuses on operations and aims to accurately model and mitigate the environmental impact of aviation, which is a joint effort by researchers from many disciplines. Accurate modeling of aircraft fuel burn, emissions, and noise is the prerequisite for the design of operational procedures and policies to mitigate negative aviation environmental impact.

In recent years, enabled by the deployment of big data technologies and the popularization of effective algorithms from applied mathematics and computer science, researchers have started to utilize data-driven analytical approaches for aviation environmental impact modeling. The application of statistics and machine learning on real-world datasets has made aviation environmental impact modeling more efficient and closer to the reality~\citep{gao2022statistics}. More specifically, methods from statistics and machine learning are employed to achieve data reduction~\citep{pagoni2017calculation, gao2021development}, efficient computation~\citep{ashok2013development, greenwood2015semiempirical}, predictive modeling~\citep{kang2018improving, baklacioglu2016modeling, khan2021prediction}, uncertainty quantification~\citep{simone2013rapid, allaire2014uncertainty, gao2019nonparametric}, pattern discovery~\citep{van2010aviation, filippone2021evaluation}, verification and validation~\citep{huynh2022delayed, simons2022comparative}, tool building~\citep{olive2019traffic, sun2019wrap}, and optimization~\citep{kim2022data}. In many typical machine learning and data mining tasks, such as classification, clustering, anomaly detection, dimensionality reduction, and information retrieval, a distance metric is required to measure the similarity or dissimilarity between objects. One such example in aviation environmental impact analysis is the problem of aircraft segmentation/grouping. There are thousands of aircraft types that are operated around the world. However, building detailed aircraft noise and performance models for each aircraft type is a long and costly process. Consequently, only a small amount of aircraft noise and performance models are available for aircraft environmental impact modeling. Using a list of aircraft features, aircraft segmentation groups aircraft types that have similar environmental impact characteristics. This practice helps select a small proportion of representative aircraft types to build aircraft noise and performance models and identify the closest ``proxy aircraft'' for the rest of the aircraft types in environmental impact modeling. 

For aircraft segmentation and other similar problems, distance metric plays a vital role in the success or failure of the task. And it is always challenging to design metrics that are well-suited to the task of interest. The majority of the studies in the literature have used Euclidean distance to compute similarity between data points or objects in a machine learning or data mining task. Other studies have considered options such as Minkowski distance and various correlation distances. A common drawback with all these options is that they assign equal weight to all aircraft features in the computation, which may pose issues in (slightly) more complex tasks. In aircraft segmentation, for example, aircraft features such as maximum takeoff weight (MTOW), wing area, and static thrust may have different influences on fuel burn, emissions, and noise. The assumption of equal significance in all aircraft features could lead to noteworthy issues. Dimensionality reduction is a potential remedy for the issues yet only applies to a fraction of the aforementioned data-driven analyses. Another solution is metric learning, which learns a tailored `weighted' distance metric for a particular task. This subfield of machine learning can potentially be beneficial to all tasks where the notion of distance metric between objects plays a crucial role~\citep{bellet2013metric}. Metric learning has found applications in fields like computer vision and bioinformatics, yet has not raised enough attention to researchers in aviation and transportation. In this work, we introduce metric learning into aircraft segmentation to make the process better reflect the actual environmental impact of aircraft. The result shows that the learned distance metrics can achieve better performance than the baselines and have good generalization properties. Overall, the three contributions of this work can be summarized as follows:
\begin{enumerate}
    \item We develop a metric learning solution to conduct aircraft segmentation for environmental impact. The proposed approach consists of representative aircraft selection, computer experiment, identification of constraints, metric learning, and evaluation methods (visual and quantitative). This work also contributes as a benchmark study of metric learning in the aviation and transportation domain.
    \item We propose a novel statistical method to identify the sets of similar and dissimilar object pairs, which are used as constraints in traditional weakly-supervised metric learning algorithms. The method can help weakly-supervised metric learning algorithms learn an effective distance metric with only a small amount of information.
    \item We demonstrate the utility of the metric learning solution through a comprehensive case study. The study involves a wide range of aircraft types, aircraft features, and representative environmental impact outputs. In the analysis of the result, we report findings and discussions from various angles.
\end{enumerate}

The remainder of the paper is organized as follows. Section~\ref{sec:litrev} provides a literature review on the topic, in both methodology and application. Section~\ref{sec:data} includes details of the data preparation step, including the selection of aircraft types, aircraft features, environmental impact outputs, and a large-scale computer experiment. Section~\ref{sec:repaircraft} describes a procedure for representative aircraft selection. Section~\ref{sec:method} introduces the general concept of metric learning, three metric learning algorithms, and identification of constraints used in this study. Section~\ref{sec:casestudy} presents a complete case study which applies the metric learning approach on aircraft segmentation for environmental impact outputs. Section~\ref{sec:remarks} discusses the limitations and extensions of the study before Section~\ref{sec:conclusion} concludes the paper.

\section{Background and Problem Formulation}\label{sec:litrev} 

\subsection{Aircraft Performance Models}

Aircraft segmentation, or aircraft grouping, has become a critical problem in aviation environmental impact modeling because of the difficulty in building Aircraft Performance Models (APMs) for a large number of aircraft types. The APM plays a central role in the environmental impact modeling of aviation. Calculations such as flight path, thrust levels, and fuel burn for terminal-area and runway-to-runway operations utilize performance models to approximate the state of an aircraft through each full air operation. The calculation results serve as primary inputs to aircraft noise and emissions estimations~\citep{gao2022representative}. In general, an APM must satisfy the following~\citep{gao2022representative,nuic2010bada}:
\begin{itemize}
    \item Can support accurate calculation of aircraft behaviours, including the kinetic, kinematic, and geometric aspects.
    \item Can cover a wide range of aircraft types, all phases of flight, and the complete operation flight envelope.
    \item Has reasonable complexity, computational properties, and maintainability.
\end{itemize}

The two primary APM specifications/databases are the Aircraft Noise and Performance (ANP) database presented by European Civil Aviation Conference (ECAC)~\citep{Doc29,Doc29-2} and EUROCONTROL’s Base of Aircraft Data (BADA) family 3~\citep{nuic2010user} and family 4~\citep{bada4}. The development of APM for an aircraft type is a rigorous process which involves close co-operations with aircraft manufacturers and operating airlines, as well as flight tests under stringent procedures. Consequently, among thousands of aircraft types that are currently or once operated, only a small fraction has APM. When there is a need to model the environmental impact of an aircraft type that is without an APM, the so-called ``missing'' aircraft type is represented by aircraft substitution. A substitute APM, also referred to as a ``proxy aircraft'', is identified through the measurement of similarity between aircraft types. Distance metric is the kernel of two relevant problems here. In the first problem, for a ``missing'' aircraft type, one must identify the closest aircraft type with APM for substitution. In the second problem, because of the limited time and/or resources in model building, people can only select a limited number of representative aircraft types to build APM, with the hope to sufficiently cover the richness and complexity of the entire aircraft population. This can be done through aircraft segmentation, which is essentially a clustering problem (more details see Section~\ref{sec:casestudy}). 

\subsection{Distance Metric}

Distance metric is a core concept in machine learning and data mining algorithms. Its formal definition and properties are given below in Definition 1.

\noindent \textbf{Definition 1} (Distance metric). Consider a set $\mathcal{X}$. A distance metric is a function $d: \mathcal{X} \times \mathcal{X} \rightarrow [0, \infty)$, such that for all $\boldsymbol{x}_i, \boldsymbol{x}_i, \boldsymbol{x}_k \in \mathcal{X}$, the following properties hold:
\begin{enumerate}
    \item Non-negativity: $d(\boldsymbol{x}_i,\boldsymbol{x}_j) \geq 0$
    \item Identity: $d(\boldsymbol{x}_i,\boldsymbol{x}_j) = 0~\Longleftrightarrow~\boldsymbol{x}_i = \boldsymbol{x}_j$
    \item Symmetry: $d(\boldsymbol{x}_i,\boldsymbol{x}_j) = d(\boldsymbol{x}_j,\boldsymbol{x}_i)$
    \item Triangle inequality: $d(\boldsymbol{x}_i,\boldsymbol{x}_j) \leq d(\boldsymbol{x}_i,\boldsymbol{x}_k) + d(\boldsymbol{x}_k,\boldsymbol{x}_j)$
\end{enumerate}

Distance metric-based machine learning and data mining techniques have been applied in many aerospace and aviation applications. Here we list some common examples in the literature. Classification has been applied to detect failure in prognostics~\citep{baptista2021classification}, analyze pilot cognitive and psychophysiological states~\citep{harrivel2016psych}, and predict aerospace structure defect~\citep{dangelo2016classification}. Clustering has been widely used to characterize and predict traffic flow patterns~\citep{murca2018identification}, extract representative flight trajectories~\citep{jensen2017development}, reduce big data from real-world operations~\citep{gao2022prem}, and study the patterns of air transportation emissions~\citep{maruhashi2022transport}. Anomaly detection has been utilized to analyze flight data and detect abnormal operations~\citep{li2015analysis}, conduct retrospective safety analysis for general aviation operations~\citep{puranik2018anomaly}, and discover the characteristics of go-around flights in commercial aviation~\citep{kumar2021classification}. Information retrieval has been used to manage and analyze aviation safety reports~\citep{tanguy2016natural}. Dimensionality reduction has been used to better visualize data and improve the performance of other supervised or unsupervised tasks~\citep{gao2022minimax}. On the use of distance metric to measure dissimilarity between data points, the most frequently used distance metrics include Euclidean distance, Manhattan distance, Cosine distance, correlation-based distance, other mixed-type distances in data mining, etc. On numeric features/attributes, Minkowski distance is a generalized form of many distance metrics used in the literature.

\noindent \textbf{Definition 2} (Minkowski distance). The Minkowski distance with order $p$ between two data points $\boldsymbol{x}, \boldsymbol{y} \in \mathbb{R}^d$ is
\begin{equation}
    d_p(\boldsymbol{x}, \boldsymbol{y}) = \left(\sum_{i=1}^d |x_i - y_i|^p\right)^{\frac{1}{p}} 
\end{equation}
where the parameter $p \in \mathbb{Z}$ could lead to different orders of proximity of points to a given data point. With $p = 1$ and $p = 2$, Minkowski distance corresponds to the Manhattan distance and the Euclidean distance, respectively. When $p \rightarrow \infty$, in the limiting case distance $d_\infty(\boldsymbol{x}, \boldsymbol{y})$ is the Chebyshev distance. The choice of $p$ is crucial for high-dimensional data analysis problems. A study~\citep{aggarwal2001surprising} shows that for a problem with fixed high dimensionality $d$, it may be preferable to use lower values of $p$, i.e., Manhattan distance is more preferable than Euclidean distance. The fractional distance metrics, where $p$ is allowed to be a fraction smaller than 1, is even more effective at preserving the meaningfulness of proximity measures in high-dimensional space for clustering and anomaly detection tasks.

Even though the Minkowski distance can provide a family of distance metrics with different properties, a notable limitation is that it assigns equal weight to all features. However, even in a slightly more complex machine learning and data mining task, such as aircraft environmental impact segmentation, different features could have different importance weights in the problem. Therefore, some sort of ``weighted distance metric'' should be applied to measure dissimilarity between aircraft types, which has gained attention in some latest works in aviation. Reference~\citep{corrado2020trajectory} utilized pre-defined weighted distance functions in an aircraft trajectory clustering problem. Reference~\citep{gao2022minimax} used weighted distance function to balance the overall influence between different data sources after data integration. Reference~\citep{gao2022multi} used LASSO parameters to quantify feature importance and a weighted Euclidean distance metric in the clustering problem.

\begin{figure}[h!]
	\centering
	\includegraphics[width=0.75\textwidth]{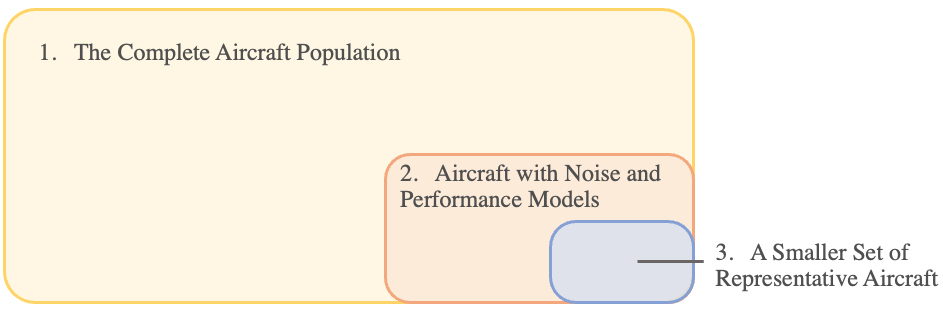}
	\caption{The three scopes of aircraft range in this work}
	\label{fig:Scopes}
\end{figure}

\subsection{Research Gap and Formulation}

Although some latest works have started to apply more sophisticated distance functions in aerospace applications, no attempt has been made to optimize the distance metric for a particular task. To fill in this gap, we develop a metric learning solution specifically for aircraft environmental impact and demonstrate its benefits over the current approach. To start with, Figure~\ref{fig:Scopes} elucidates the three scopes of aircraft type that are relevant to this study. The first scope (scope 1) is the complete aircraft type population. Thousands of different aircraft types have been operated around the world over the past decades. For example, in Federal Aviation Administration (FAA)'s Aviation Environmental Design Tool (AEDT)~\citep{lee2020aviation} database, there are currently over 3,000 aircraft types. Even though not all of them are `active' aircraft types involved in recent environmental impact modeling efforts, the overall aircraft population in need of environmental impact analysis is still large. The second scope (scope 2) includes aircraft types that have APM. Only a (small) subset of the aircraft population has APM because of the high cost and time of building one. For example, less than 300 aircraft types currently have ANP model -- a commonly used APM. Because APM is essential for the computation of aircraft environmental impact, scope 2 is the subset of aircraft types that have the `ground truth' results on environmental impact. The ultimate objective is to learn tailored distance metrics from the partial information in scope 2 and apply them on the entire population in scope 1. However, since the rest of aircraft types in scope 1 do not have `ground truth' environmental impact results, we cannot test how well the learned distance metrics can generalize to a larger population. Therefore, within scope 2, we select a subset of representative aircraft types and form the third scope (scope 3), learn tailored distance metrics from it, and test them with the `ground truth' results in scope 2. A good generalization property from scope 3 to scope 2 is an indicator that the metric learning solution has the potential to perform well when generalizing from scope 2 to scope 1.

\begin{figure}[h!]
	\centering
	\includegraphics[width=0.91\textwidth]{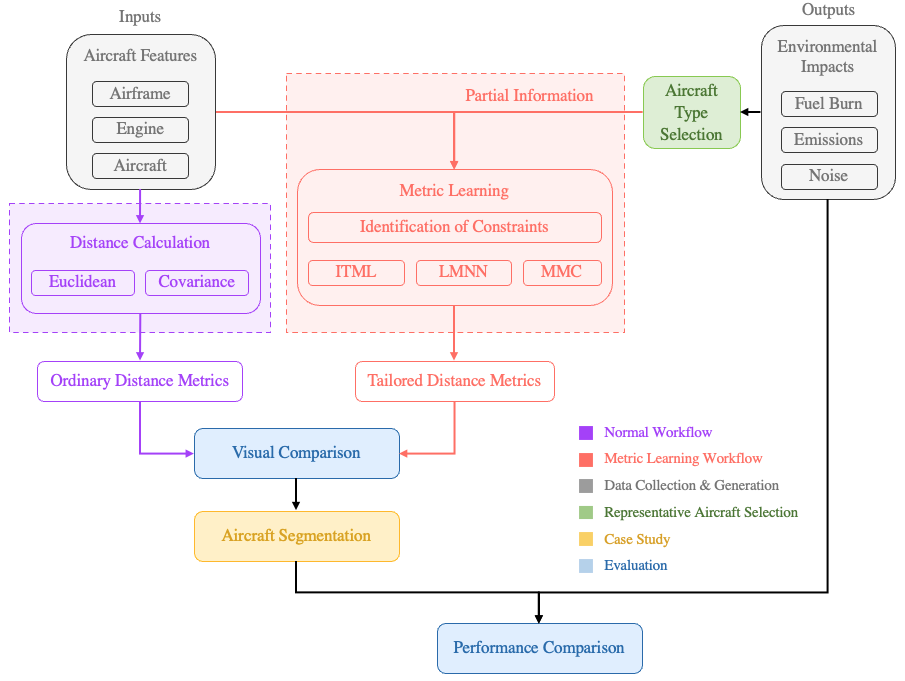}
	\caption{The metric learning formulation for aircraft environmental impact analysis}
	\label{fig:Approach}
\end{figure}

Figure~\ref{fig:Approach} depicts the formulation of this work. We first make a distinction between inputs and outputs in this process. The inputs are a set of aircraft features that every aircraft type in the complete population (scope 1 in Figure~\ref{fig:Scopes}) must have. In the actual aircraft segmentation, one uses the inputs and a distance metric to determine if two aircraft types are similar enough. In machine learning, such an analysis purely based on the inputs is called unsupervised learning. The outputs are aircraft environment impact including fuel burn, emissions, and noise. Because only aircraft types with APM have known output values, only partial information is available on the outputs. Now, we can better utilize these information and make the process semi-(or weakly-) supervised through distance metrics informed by partial information. The two different workflows, conventional workflow and metric learning workflow, are colored in purple and red, respectively in Figure~\ref{fig:Approach}. The conventional workflow is unsupervised and uses input information only; the metric learning workflow is weakly-supervised and also utilize the partial output information. The resulting distance metrics from both workflows will be used in visual comparison and the aircraft segmentation case study. For better testing purposes, a representative aircraft type selection step is taken to form a subset of aircraft (scope 3 in Figure~\ref{fig:Scopes}), whose environmental impact outputs will be used as partial information in metric learning. Finally, several quantitative measures will be employed to evaluate the performance of different distance metrics in the aircraft environmental impact segmentation task.

\section{Data preparation}\label{sec:data} 

The dataset used in the case study is from two main sources. On the input side, the aircraft features are collected and integrated from multiple tables from the AEDT FLEET database and additional feature engineering efforts. On the output side, the environmental impact outputs of aircraft are obtained through running a large-scale computer experiment in AEDT. This section provides justifications on the selection of aircraft features and environmental impact outputs, as well as the setting of the computer experiment. 

\subsection{Aircraft Features}

In this study, we focus on the environmental impact of fixed-wing aircraft. The set of aircraft features should satisfies two conditions: (1) it comprehensively represents an aircraft's characteristics that are potentially related to its environmental impact; and (2) it is available for all aircraft types in the population. Since each aircraft type is a unique combination of airframe and engine, features from three data sources are first considered: airframe, engine, and aircraft (airframe + engine combination). Table~\ref{tbl:ori-feature} in Appendix includes a complete list of the 36 selected aircraft features from these three sources. The fuel burn and emissions indices are part of the engine features, and here we treat this special group as the fourth category. There also exists different attribute types, such as numeric, nominal, and ordinal attributes. Details on the levels and descriptions of the categorical features can be found in Table~\ref{tbl:cat-feature} in Appendix. In addition, we use feature engineering and engineering knowledge in aircraft design to create additional features that may represent crucial nonlinear feature interactions. The inclusion of these additional engineered features can increase the predictive power of the feature set and simplify statistical inference, because the interaction information can be incorporated into simpler models with better interpretation. The 10 additional aircraft features created from feature engineering and their formulations can be found in Table~\ref{tbl:eng-feature} in Appendix.

In the actual analysis, one-hot encoding is used to further represent categorical features. For a specific categorical feature, one-hot encoding creates a binary column (or new feature) for each category. For instance, the feature ``Location of Engines (ENGINE\_LOCATION)'' has three categories -- Fuselage/Tail (F), Internal (I), and Wing (W), which is then converted to a total of three new features, each indicates a True (1) or False (0) on an individual category. This further increases the total number of aircraft features to 66 in this study. All numeric features are standardized such that they have zero mean and unit variance.

\subsection{Large-scale Computer Experiment} 

Because of the infeasibility of actually collecting fuel burn, emissions, and noise results for a wide range of aircraft types under the same, standard conditions, in this study the environmental impact outputs are obtained through computer simulations via AEDT. Table~\ref{tbl:comp-sim} shows the factors and their settings in this computer experiment. On the aircraft type selection, we attempt to include all aircraft types in the current ANP database. Another consideration for aircraft selection is whether an aircraft type has available data on the selected features -- mainly the features in Table~\ref{tbl:ori-feature} in Appendix. After the data integration process, this large-scale computer experiment includes 214 unique aircraft types. A complete list of the 214 selected aircraft types, which forms the scope 2 in Figure~\ref{fig:Scopes}, is given in Table~\ref{tbl:aircraft} in Appendix. This range of aircraft types comprehensively covers different classes of aircraft, such as regional jet (RJ), large single aisle (LSA), small twin aisle (STA), large twin aisle (LTA), general aviation (GA) aircraft, and some military aircraft. It also includes all mainstream aircraft models that are currently operated around the world.

\begin{table}[htbp]
\centering
\caption{Factors and settings in the computer experiment}
\label{tbl:comp-sim}
\begin{tabular}{c|c}
\hline
\textbf{Factors} & \textbf{Details}                                        \\ \hline
Aircraft         & 214 different aircraft types                            \\ 
Stage Length     & Maximum stage length for all aircraft                   \\ 
Operation       & Departure                                  \\ 
Profile          & Standard profile                                        \\ 
Airport          & Hartsfield-Jackson Atlanta International Airport (KATL) \\ 
Runway           & 08L -- 26R                                               \\ 
Weather          & Standard day weather                                    \\ 
Receptor Set     & 301*121, spacing 0.2 n.mi                               \\ \hline
\end{tabular}
\end{table}

Most supervised machine learning or metric learning techniques require that only one value is used to represent each output. Therefore, we choose one representative setting for running the computer experiment and calculating each aircraft's environmental impact. Stage length is a term in aviation used to describe the distance of a single flight from takeoff to landing. Each aircraft type has a discrete number of stage lengths, which ranges from 1 to 10, depending on the aircraft type. If an aircraft type has multiple stage lengths, a higher stage length is associated with a larger takeoff weight. This experiment uses the maximum stage length to represent each aircraft type in environmental impact modeling because it represents an aircraft's `capacity' in all aspects. On the operation side, the experiment runs departure operation because of its main role in terminal area fuel burn, emissions, and community noise impact. A departure profile is a sequence of steps that a simulation environment follows to fly the aircraft in computer simulation~\citep{gao2021development}. For example, a representative departure profile consists of a takeoff step, a constant speed climb step to a safe altitude, a thrust cutback step, and a final climb-out step to 10,000 ft. The standard profile in AEDT is used in the simulation. All cases are run at the runway 08L -- 26R of Hartsfield-Jackson Atlanta International Airport (KATL) under standard day weather conditions. KTAL is chosen because it has a moderate elevation at 1,026 ft and long enough runways for all aircraft types. To measure the noise exposure, this experiment uses a high-fidelity receptor grid which covers an area of 60.2 nautical miles by 24.2 nautical miles.

After the computer simulations and a post-calculation process, the experiment generates, for each aircraft type, a single value for each environmental impact output. Admittedly, variability exists in factors such as stage length, airport, and weather. The impacts of these variabilities are worthy of further investigations and will be further discussed in Section~\ref{sec:remarks}. At the moment, a fundamental assumption here is that the result of the experiment described in Table~\ref{tbl:aircraft} is representative of the relative environmental impact characteristics between different aircraft types (i.e., each output is at around the mode, or most likely status), which suffices for a credible metric learning practice.

\subsection{Environmental Impact Outputs}

The outputs of interest in this study are aircraft environmental impact outputs in the terminal area. The range of the environmental impact outputs must properly cover all of fuel burn, emissions, and noise. A list of the 6 most significant environmental impact outputs used in relevant studies is given in Table~\ref{tbl:outputs}. The emissions outputs mainly include nitrogen oxides (NO$_x$), carbon monoxide (CO), and non-volatile particulate matter (nvPM). The noise outputs are mainly noise contour areas at difference noise levels. For departure, the Sound Exposure Level (SEL) 80 dB and 90 dB contours are measured. A computational code~\citep{gao2021development} has been developed by the author of this study to process the output files from AEDT and perform computations. The code has undergone a rigorous verification and validation (V\&V) process to ensure that it outputs accurate values on all environmental impact outputs.

\begin{table}[htbp]
\centering
\caption{The list of 12 environmental impact outputs}
\label{tbl:outputs}
\begin{tabular}{ccccc}
\hline
\textbf{No.} & \textbf{Category}           & \textbf{Code}        & \textbf{Output Name}                    & \textbf{Unit} \\ \hline
1            & \multirow{1}{*}{Fuel Burn}  & D\_Fuel              & Departure Fuel Burn                     & kg             \\ \hline
2            & \multirow{3}{*}{Emissions} & D\_NOX               & Departure NOx Emissions                 & kg             \\ 
3            &                             & D\_CO                & Departure CO Emissions                  & kg             \\ 
4           &                             & D\_NVPM        & Departure nvPM Emissions & kg             \\ \hline
5           & \multirow{2}{*}{Noise}     & D\_Noise\_80\_Area   & Departure SEL Contour Area at 80 dB     & n.mi.$^2$     \\ 
6           &                             & D\_Noise\_90\_Area   & Departure SEL Contour Area at 90 dB     & n.mi.$^2$     \\ 
\hline
\end{tabular}
\end{table}

\section{Representative Aircraft Selection}\label{sec:repaircraft}

The selection of representative aircraft types has two levels of significance in this work. First, as discussed in Section~\ref{sec:litrev}, to validate the metric learning approach, we must select a subset of the aircraft with APM to learn tailored distance metrics for environmental impact outputs. This is to test whether the distance metrics learned from partial information can apply and generalize to a larger population. Second, because of the limited time and resources, practitioners can only select a subset of representative aircraft types for building detailed APM. The objective is that the selected aircraft types can sufficiently cover the richness and complexity of the entire aircraft population.

In this part we aim to select $k$ representative aircraft types, also referred to as ``prototypes'', from the population with $n$ aircraft types to sufficiently cover the population by attaining certain optimality. Among the multiple criteria of representativeness in the literature, we use the minimax consideration as the primary strategy for selecting prototypes from the population. Minimax designs have been applied on experimental design~\citep{johnson1990minimax,tan2013minimax} and application cases such as air monitoring~\citep{mak2018minimax} and optimal placement of sensors~\citep{vanli2012minimax}. In a minimax selection problem, we seek a set of prototypes that minimizes the maximum distance (dissimilarity) from any object in the population to its nearest prototype~\citep{gao2022minimax}. This ensures that at each budget level $k$, every aircraft in the population can be maximally covered by the selected subset. In a formal definition, let $\mathcal{D} = \{\{\boldsymbol{x}_i\}_{i=1}^n: \boldsymbol{x}_i \in \mathbb{R}^p \}$ be the complete population in a $p$-dimensional space, a set of minimax prototypes $\mathcal{P}^* = \{\{\boldsymbol{p}_i\}_{i=1}^k: \boldsymbol{p}_i \in \mathcal{D} \}$ is defined as the solution of
\begin{equation}\label{eqn:mimimax}
\displaystyle \mathcal{P}^* \leftarrow \underset{\mathcal{P} \in \mathcal{D}}{\mathrm{argmin}} \sup_{\boldsymbol{x} \in \mathcal{D}} \| \boldsymbol{x}-\mathcal{C}(\boldsymbol{x},\mathcal{P})\|
\end{equation}
where $\mathcal{C}(\boldsymbol{x},\mathcal{P}) \leftarrow \mathrm{argmin}_{\boldsymbol{z} \in \mathcal{P}} \|\boldsymbol{x}-\boldsymbol{z}\|$ is the closest prototype to $\boldsymbol{x}$ under norm $\| \cdot \|$. Finding an exact solution to the minimax problem is NP-hard. Therefore, we consider two machine learning-based approaches for the approximate solutions to the minimax problem. The two methods, both based on clustering, are hierarchical clustering with minimax linkage~\citep{bien2011hierarchical}, and minimax k-med~\citep{gao2022minimax}. We choose hierarchical clustering with minimax linkage because of the deterministic result and theoretical properties. 

Agglomerative hierarchical clustering groups similar objects together using a bottom-up approach and represents the sequence of clusterings as a dendrogram~\citep{bien2011hierarchical}. Applying on the aircraft population, it starts with $n$ singleton aircraft clusters at the leaves and merges two closest clusters at each stage until only one cluster remains at the root. The distance measure between clusters, referred to as linkage, is an important choice in hierarchical clustering. The minimax linkage is a concept initially proposed by genetics researchers~\citep{ao2004clustag}. For any object (aircraft) $\boldsymbol{x}$ and cluster $C$, define
\begin{equation}
d_\text{max}(\boldsymbol{x},C) = \max_{\boldsymbol{x}' \in C} d(\boldsymbol{x},\boldsymbol{x}') 
\end{equation}
as the distance from $\boldsymbol{x}$ to its farthest object in $C$. Then, define the maximum radius of cluster $C$ as
\begin{equation}
r(C) = \min_{\boldsymbol{x} \in C} d_\text{max}(\boldsymbol{x},C)
\end{equation}
which is the $d_\text{max}$ of the object in $C$ whose farthest object is closest. The corresponding minimizing object is also called the \textit{prototype} of $C$, at which a closed ball of radius $r(C)$ covers all of $C$. Finally, the minimax linkage between two clusters $G$ and $H$ measures the inter-cluster distance by the minimax radius of the resulting merged cluster, defined as
\begin{equation}
d(G,H) = r(G \cup H)
\end{equation}
where each newly merged cluster $G \cup H$ is naturally represented by an associated prototype, which is its most `central' object. This set of selected prototypes enjoys a number of desirable theoretical properties, of which the most crucial property is the \textit{maximum distortion guarantee}. In agglomerative hierarchical clustering, the dendrogram places each leaf (object) at height 0 and every interior node (cluster merge) at a height that is equal to the distance between the merged clusters, i.e., $h(G \cup H) = d(G,H)$. The maximum distortion guarantee states that, with the minimax linkage, cutting the dendrogram at height $h$ yields a set of clusters $C_1,...,C_k$ and associated prototypes $\boldsymbol{p}_1,...,\boldsymbol{p}_k$ such that for every cluster $C_i$, we have $d_\text{max}(\boldsymbol{p}_i,C_i) \leq h$. In other words, all objects in the population are within distance $h$ of its closet prototype, which is similar to the famous set cover problem. The maximum distortion guarantee ensures that very aircraft type in the population is sufficiently covered and represented by the selected prototypes.

\section{Metric Learning Methodologies}\label{sec:method}

This section introduces the metric learning techniques in detail. The section begins with a brief introduction on the background and essential elements of metric learning, followed by the detailed formulations of three selected metric learning methods that will be tested in the case study. The last subsection covers how different constraints for metric learning algorithms are identified in this study, including a novel statistical approach to effectively identify the sets of similar and dissimilar objects. 

\subsection{Preliminaries of Metric Learning}

\begin{figure}[h!]
	\centering
	\includegraphics[width=0.75\textwidth]{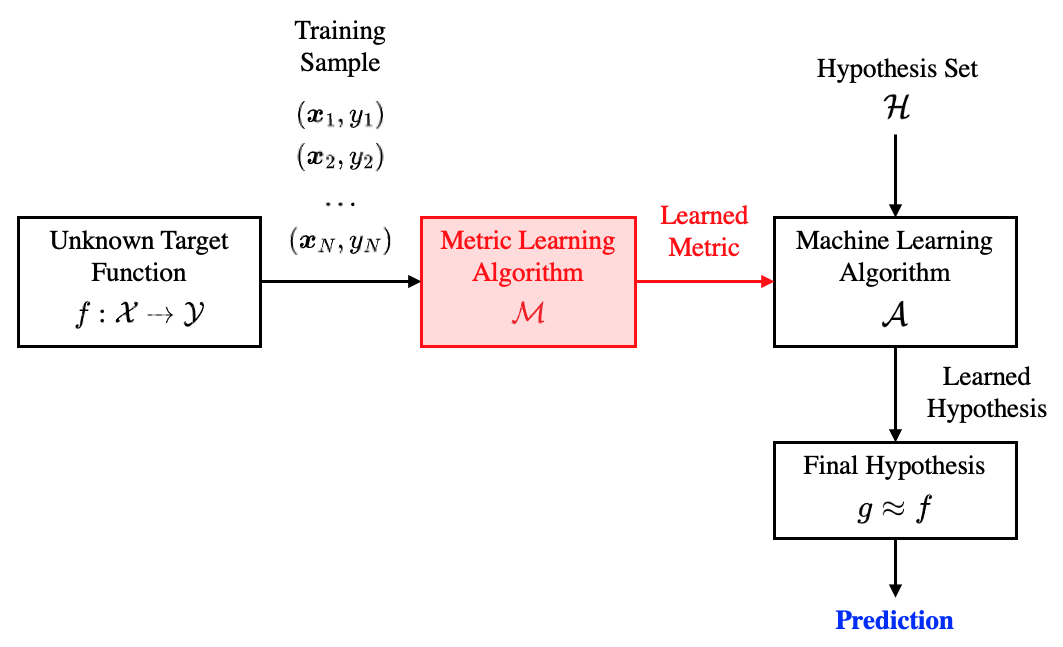}
	\caption{The role of metric learning in a machine learning problem}
	\label{fig:MetricLearningProcess}
\end{figure}

A metric learning problem focuses on learning a tailored distance metric for a particular task~\citep{kulis2013metric}, and is useful for machine learning or data mining techniques that rely on the computation of distances or similarities over pairs of objects (for example, clustering and classification). It addresses the question \textit{how to properly assess the distance or similarity between the pairs of objects}, which arises as a challenge in complex tasks. Most metric learning algorithms learn task-specific distance functions in a supervised manner. Figure~\ref{fig:MetricLearningProcess} depicts a common metric learning process, which includes an additional metric learning step compared to a traditional machine learning problem. With a learned distance metric from training sample, the objective is to perform better than a predictor induced by a standard metric~\citep{bellet2013metric}. The Mahalanobis distance, a distance function used in metric learning, is a pairwise real-valued metric function originally proposed by~\citep{mahalanobis1936generalized}.

\noindent \textbf{Definition 3} (Mahalanobis distance). Consider a $d$-dimensional set $\mathcal{X}$. Let two clouds of data points $\mathcal{X}_1, \mathcal{X}_2 \in \mathcal{X}$, the Mahalanobis distance between the two points $\boldsymbol{x}_i \in \mathcal{X}_1$ and $\boldsymbol{x}_j \in \mathcal{X}_2$ is
\begin{equation}\label{eqn:md}
    d_\Sigma(\boldsymbol{x}_i, \boldsymbol{x}_j) = \sqrt{(\boldsymbol{x}_i - \boldsymbol{x}_j)^\top \Sigma^{-1} (\boldsymbol{x}_i - \boldsymbol{x}_j)} 
\end{equation}
where $\Sigma \in \mathbb{R}^{d \times d}$ is the covariance matrix of data in sets $\mathcal{X}_1$ and $\mathcal{X}_2$. When computing the distance between a data point and a data cloud (or distribution), the Euclidean distance ignores the scatter/variance of the cloud which could miss vital information in the distance measure. The Mahalanobis distance considers the variance of data and performs better in cases where the properties of data distribution must be taken into account.

\noindent \textbf{Definition 4} (Generalized Mahalanobis distance). In Mahalanobis distance, the covariance matrix $\Sigma$ and its inverse $\Sigma^{-1}$ are positive semi-definite (PSD). When $\Sigma^{-1}$ is replaced by a PSD weight matrix $M \succeq 0$, we have the generalized Mahalanobis distance between the two points
\begin{equation}
    d_M(\boldsymbol{x}_i, \boldsymbol{x}_j) = \sqrt{(\boldsymbol{x}_i - \boldsymbol{x}_j)^\top M (\boldsymbol{x}_i - \boldsymbol{x}_j)} 
\end{equation}

It can be shown that the generalized Mahalanobis distance is a valid distance metric~\citep{ghojogh2022metric}. The matrix $M$ can now weight the dimensions when computing the distance. Therefore, learning a proper matrix $M$ is a major component of Mahalanobis distance metric learning. When $W = I$, the Euclidean distance is a special case of the Mahalanobis distance. Most methods learn the metric (essentially the PSD matrix $M$) in a weakly supervised manner~\citep{bellet2013metric}. In a dataset, some data points are similar in some sense, e.g., they have similar patterns or characteristics. In contrast, there also exists dissimilar data points. Here we use $\mathcal{S}$ and $\mathcal{D}$ to denote the sets of similar pair points and dissimilar
pair points (or positive/negative pairs) respectively:
\begin{equation}
    \begin{aligned}
    & (\boldsymbol{x}_i, \boldsymbol{x}_j) \in \mathcal{S} \quad \text{if $\boldsymbol{x}_i$ and $\boldsymbol{x}_j$ are similar}\\
    & (\boldsymbol{x}_i, \boldsymbol{x}_j) \in \mathcal{D} \quad \text{if $\boldsymbol{x}_i$ and $\boldsymbol{x}_j$ are dissimilar}
    \end{aligned}
\end{equation}

For example, if class labels are available, then $(\boldsymbol{x}_i, \boldsymbol{x}_j) \in \mathcal{S}$ if $\boldsymbol{x}_i$ and $\boldsymbol{x}_j$ are in the same class; $(\boldsymbol{x}_i, \boldsymbol{x}_j) \in \mathcal{D}$ if $\boldsymbol{x}_i$ and $\boldsymbol{x}_j$ are in different classes. Other than the pair constraints, the triplet-based (or relative) constraints have the form
\begin{equation}
    (\boldsymbol{x}_i, \boldsymbol{x}_j, \boldsymbol{x}_k) \in \mathcal{R} \quad \text{if $\boldsymbol{x}_i$ is more similar to $\boldsymbol{x}_j$ than to $\boldsymbol{x}_k$}
\end{equation}

A metric learning algorithm aims to learn the weight matrix $M$ that best agrees with the constraints in $\mathcal{S}$, $\mathcal{D}$, and $\mathcal{R}$ so that the distances of similar points become smaller and the distances of dissimilar points become larger. The optimization problem for metric learning typically has the following general form
\begin{equation}\label{eqn:mlgeneral}
    \min_M \quad L(M, \mathcal{S}, \mathcal{D}, \mathcal{R}) + \lambda R(M) 
\end{equation}
where $L(M, \mathcal{S}, \mathcal{D}, \mathcal{R})$ is a loss function which incurs a penalty when constraints are violated, $R(M)$ is some regularizer on matrix $M$ and $\lambda \geq 0$ is the regularization parameter. Different metric learning algorithms differ by the choices of metric, constraints, loss function, and regularizer. Most metric learning algorithms follow the general form in Equation~\ref{eqn:mlgeneral}.

Since the early metric learning algorithms such as Mahalanobis Metric for Clustering (MMC)~\citep{xing2002metric} and Schultz \& Joachims~\citep{Schultz2003distance}, many state-of-the-art algorithms have emerged. Each metric learning algorithm has its intrinsic properties, such as the angle of problem solving, type of metric, ability to leverage unsupervised data, ability to reduce dimensionality, optimality of the solution, etc. In this work, we apply three classic metric learning algorithms for aircraft environmental impact segmentation. They are MMC, Large Margin Nearest Neighbors (LMNN)~\citep{weinberger2009distance} and Information-Theoretic Metric Learning (ITML)~\citep{davis2007information}. They are chosen in this study for the following considerations:
\begin{enumerate}
    \item \textbf{Optimality of the solution}: in this study we select metric learning methods which usually have convex formulations (global optimality) and is less prone to overfitting.
    \item \textbf{Global metric learning}: at the current stage, for each environmental impact output we hope to learn one global distance metric that can be applied to all aircraft types.
    \item \textbf{Different angles}: the methods should provide different problem solving angles. For example, MMC is a classic and interpretable method that utilizes side information; ITML is an information-theoretic approach; LMNN is a nearest-neighbors approach. Their constraints also have different forms.
\end{enumerate}

\subsection{Unsupervised Metric Learning}

An unsupervised metric learning method generates distance metrics without utilizing any information from the outputs. While the main metric learning methods used in this work are weakly-supervised and utilize partial information from the outputs, we also apply an unsupervised method and use it as a baseline in the performance comparison. The unsupervised method computes the covariance matrix of the inputs (aircraft features) and use it in the Mahalanobis distance in Equation~\ref{eqn:md}.

\subsection{Mahalanobis Metric for Clustering (MMC)}

As the earliest metric learning algorithm, MMC is a weakly-supervised learning approach which utilizes ``side information'' -- whether certain pairs of data points are similar or dissimilar. The optimization problem of MMC has a convex formulation and minimizes the sum of distances between similar points while keeping the sum of distances between dissimilar examples above certain threshold:
\begin{equation}\label{eqn:oriMMC}
\begin{aligned}
\min_{M} \quad & \sum_{(\boldsymbol{x}_i, \boldsymbol{x}_j) \in \mathcal{S}} d_M^2(\boldsymbol{x}_i, \boldsymbol{x}_j)\\
\text{subject to} \quad & \sum_{(\boldsymbol{x}_i, \boldsymbol{x}_j) \in \mathcal{D}} d_M(\boldsymbol{x}_i, \boldsymbol{x}_j) \geq 1\\
& M \succeq 0
\end{aligned}
\end{equation}

This optimization problem is solved by a projected gradient approach which involves the full eigenvalue decomposition of $M$ at each iteration. The choice of constant 1 in the constraint is arbitrary but not important, because replacing it with another positive constant $c$ just results in $M$ being replaced by $c^2 M$. A nice property of this method is that, to learn a diagonal matrix $M = \text{diag}(M_{11}, M_{22}, ..., M_{nn}$), one can use the Newton-Raphson method to efficiently solve the problem. It can be shown that the original problem in Equation~\ref{eqn:oriMMC} is equivalent to 
\begin{equation}
\begin{aligned}
\min_{M} \quad & \sum_{(\boldsymbol{x}_i, \boldsymbol{x}_j) \in \mathcal{S}} d_M^2(\boldsymbol{x}_i, \boldsymbol{x}_j) - \log \left( \sum_{(\boldsymbol{x}_i, \boldsymbol{x}_j) \in \mathcal{D}} d_M(\boldsymbol{x}_i, \boldsymbol{x}_j) \right)\\
\text{subject to} \quad & M \succeq 0\\
& M = \text{diag}(M_{11}, M_{22}, ..., M_{nn}) \quad \text{(Optional)}
\end{aligned}
\end{equation}

Restricting $M$ to be diagonal is optional in MMC and may have an important implication on feature selection, depending on how the features are normalized. MMC is then essentially learning a distance metric which assigns different weights to different features. Consequently, $d_M(\boldsymbol{x}_i, \boldsymbol{x}_j)$ becomes a weighted Euclidean distance function. Ideally, this generates insights on which features are more influential on a certain output, as a higher weight indicates that the corresponding feature is more influential in the problem. An advantage of $M$ being diagonal is the interpretability of the result, although it may be accompanied by a loss of generalizability compared to the full Mahalanobis distance learning. In our case study, we restrict $M$ in MMC to be diagonal.

\subsection{Information-Theoretic Metric Learning (ITML)}

Information-Theoretic Metric Learning (ITML)~\citep{davis2007information} applies a special information-theoretic regularizer to the problem. The core idea is to regularize the Mahalanobis matrix $M$ to be as close as possible to a given matrix $M_0$. An natural information-theoretic approach is applied to quantify the ``closeness'' between matrices $M$ and $M_0$. Consider a multivariate Gaussian distribution parameterized by mean $\boldsymbol{\mu}$ and matrix $M$, $p(\boldsymbol{x};M) = (1/Z) \exp(-d_M(\boldsymbol{x},\boldsymbol{\mu}))$, where $Z$ is a normalizing constant, then the distance between two multivariate
Gaussians of the same mean and matrices $M$ and $M_0$ can be measured by the following the Kullback–Leibler (K-L) divergence:
\begin{equation}\label{eqn:KL}
    \text{KL} \left(p(\boldsymbol{x};M_0)\|p(\boldsymbol{x};M)\right) = \int p(\boldsymbol{x};M_0) \log \frac{p(\boldsymbol{x};M_0)}{p(\boldsymbol{x};M)} d \boldsymbol{x}
\end{equation}

The K-L divergence ``between two matrices'' can be expressed as a particular type of Bregman divergence -- the LogDet divergence. The LogDet divergence between two positive definite matrices is defined as
\begin{equation}\label{eqn:logdet}
    D_{ld}(M,M_0) = \text{tr} (M M_0^{-1}) - \log\det (M M_0^{-1}) - d
\end{equation}
where $d$ is the dimension of the problem (number of features). As for the relationship between Equations~\ref{eqn:KL} and~\ref{eqn:logdet}, through derivations it can be shown that
\begin{equation}
    \text{KL} \left(p(\boldsymbol{x};M_0)\|p(\boldsymbol{x};M)\right) = \frac{1}{2} D_{ld}(M,M_0)
\end{equation}

In ITML, $M_0$ is a positive definite matrix which we want $M$ to remain close to. One common choice of $M_0$ is the identity matrix $I$, under which the learned Mahalanobis distance is regularized to the Euclidean distance. Minimizing $D_{ld}(M,I)$ provides a natural way of preserving the positive semi-definiteness of $M$. On the constraints side, ITML uses an upper bound $u$ between similar data points and a lower bound $l$ between dissimilar data points. With the slack variables, the optimization problem is formulated as
\begin{equation}
\begin{aligned}
\min_{M} \quad & \text{tr} (M) - \log\det (M) + \gamma \sum_{i,j} \xi_{ij}\\
\text{subject to} \quad & d_M^2(\boldsymbol{x}_i, \boldsymbol{x}_j) \leq \mu + \xi_{ij} \quad \forall (\boldsymbol{x}_i, \boldsymbol{x}_j) \in \mathcal{S}\\
& d_M^2(\boldsymbol{x}_i, \boldsymbol{x}_j) \geq v - \xi_{ij} \quad \forall (\boldsymbol{x}_i, \boldsymbol{x}_j) \in \mathcal{D}\\
\end{aligned}
\end{equation}
where $\gamma$ is a trade-off parameter that can be optimized via cross-validation in the training process. The LogDet divergence is also referred to Stein's loss in the statistics community. ITML is the first method which introduces LogDet divergence regularization into Mahalanobis distance learning and provides a different angle to tackle the problem. 

\subsection{Large Margin Nearest Neighbors (LMNN)}

Large Margin Nearest Neighbors (LMNN)~\citep{weinberger2009distance,weinberger2008fast} is one of the most widely-used methods for metric learning. It is an approach driven by nearest neighbors. LMNN is based on two intuitions that the learned distance metric should satisfy, as each data point should: (1) belong to the same class as its $k$ nearest neighbors, also referred to as target neighbors, and (2) be widely separated from instances of other classes, also referred to as imposters. To increase the robustness of $k$-NN classification, LMNN aims to maintain a finite safety margin between imposters and the perimeters established by target neighbors. The loss function of LMNN consists of two competing terms. The first term ``pull'' target neighbors closer together, and the second term ``push'' examples from different classes further apart. From another angle, one term penalizes large distances between target neighbors, while the other term penalizes small distances between differently labeled examples. Mathematically, the two terms can be expressed by
\begin{equation}\label{eqn:pillpush}
    \begin{aligned}
    \epsilon_{\text{pull}} (M) &= \sum_{(\boldsymbol{x}_i, \boldsymbol{x}_j) \in \mathcal{S}} d_M^2(\boldsymbol{x}_i, \boldsymbol{x}_j)\\
    \epsilon_{\text{push}} (M) &= \sum_{(\boldsymbol{x}_i, \boldsymbol{x}_j, \boldsymbol{x}_k) \in \mathcal{R}} \left[1+d_M^2(\boldsymbol{x}_i, \boldsymbol{x}_j) - d_M^2(\boldsymbol{x}_i, \boldsymbol{x}_k)\right]_+
    \end{aligned}
\end{equation}
where $[z]_+ = \max(z,0)$ denotes the hinge loss which makes no contribution to the term with a negative argument. And like the MMC formulation in Equation~\ref{eqn:oriMMC}, the choice of unit margin in LMNN is an arbitrary convention that is less critical. The two competing terms are then combined into a single loss function
\begin{equation}\label{eqn:pillpushcombined}
    \epsilon(M) = (1-\mu) \epsilon_{\text{pull}} (M) + \mu \epsilon_{\text{push}} (M)
\end{equation}
where $\mu \in [0,1]$ is a weighting parameter which controls the ``pull/push'' trade-off. The value of $\mu$ will be selected via cross-validation, although the authors of LMNN suggest that $\mu = 0.5$ works well in general~\citep{weinberger2009distance}. Substituting Equation~\ref{eqn:pillpush} into Equation~\ref{eqn:pillpushcombined} gives us the loss function for LMNN:
\begin{equation}\label{eqn:LMNNobj}
    \epsilon(M) = (1-\mu) \sum_{(\boldsymbol{x}_i, \boldsymbol{x}_j) \in \mathcal{S}} d_M^2(\boldsymbol{x}_i, \boldsymbol{x}_j) + \mu \sum_{(\boldsymbol{x}_i, \boldsymbol{x}_j, \boldsymbol{x}_k) \in \mathcal{R}} \left[1+d_M^2(\boldsymbol{x}_i, \boldsymbol{x}_j) - d_M^2(\boldsymbol{x}_i, \boldsymbol{x}_k)\right]_+
\end{equation}
which a piecewise linear, convex function of the elements in matrix $M$. The optimization problem of LMNN is formulated as a semi-definite program (SDP)~\citep{weinberger2009distance}, which can be efficiently solved. To convert Equation~\ref{eqn:LMNNobj} into a standard form of SDP, slack variables $\xi_{ijk} \geq 0$ are introduced to mimic and replace the hinge loss. Finally, using slack variables the optimization problem of LMNN is formulated as the following SDP:
\begin{equation}
\begin{aligned}
\min_{M} \quad & (1-\mu) \sum_{(\boldsymbol{x}_i, \boldsymbol{x}_j) \in \mathcal{S}} d_M^2(\boldsymbol{x}_i, \boldsymbol{x}_j) + \mu \sum_{i,j,k} \xi_{ijk}\\
\text{subject to} \quad & d_M^2(\boldsymbol{x}_i, \boldsymbol{x}_k) - d_M^2(\boldsymbol{x}_i, \boldsymbol{x}_j) \geq 1 - \xi_{ijk} \quad \forall (\boldsymbol{x}_i, \boldsymbol{x}_j, \boldsymbol{x}_k) \in \mathcal{R}\\
& \xi_{ijk} \geq 0\\
& M \succeq 0
\end{aligned}
\end{equation}

An advantage of LMNN is that the constraints are defined in a local way, i.e., the $k$ nearest neighbors of any data point form the target neighbors and belong to the same class while the other data points are considered as imposters from other classes. Because of this, LMNN has a wide applicability in many different problems. 

\subsection{Identification of Constraints}

\begin{figure}[h!]
	\centering
        \includegraphics[width=0.3\textwidth]{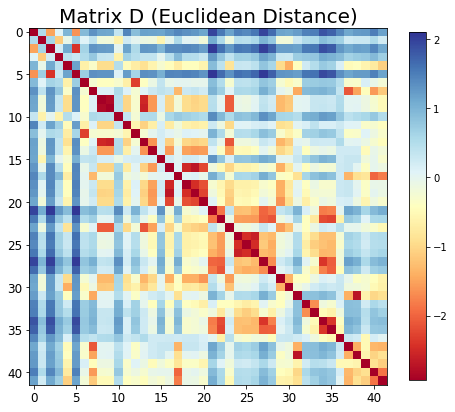}
        \hspace{0.2cm}
        \includegraphics[width=0.305\textwidth]{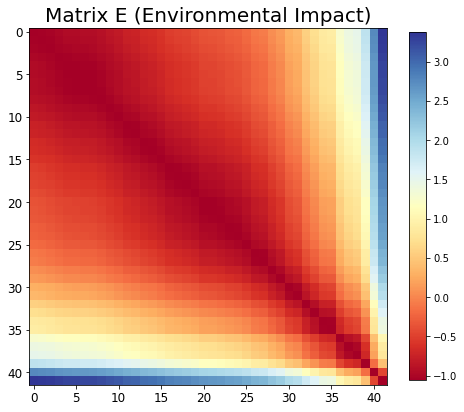}
        \hspace{0.2cm}
        \includegraphics[width=0.3\textwidth]{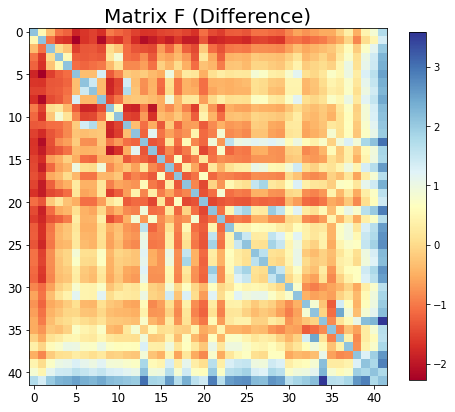}\\
        \vspace{0.25cm}
        \includegraphics[width=0.3\textwidth]{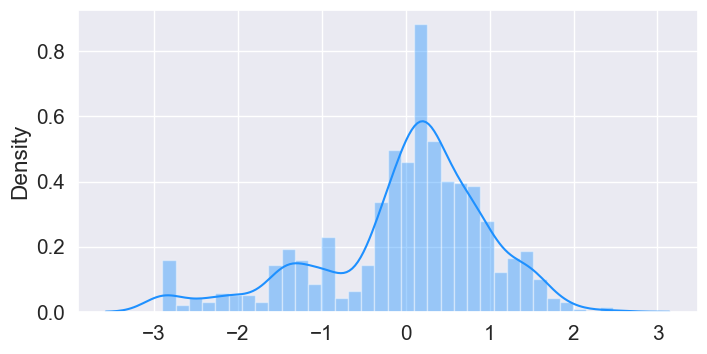}
        \hspace{0.2cm}
        \includegraphics[width=0.3\textwidth]{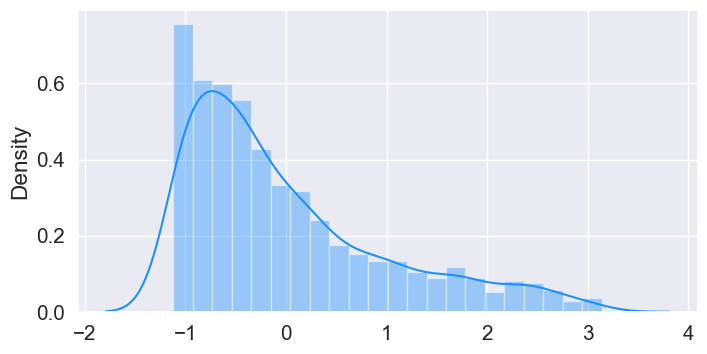}
        \hspace{0.2cm}
        \includegraphics[width=0.3\textwidth]{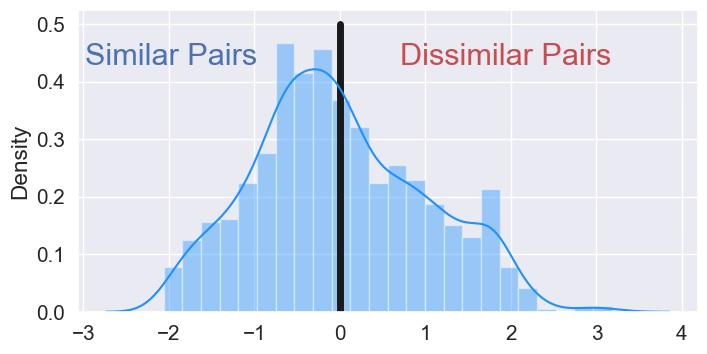}
	\caption{The three matrices used in the identification of similar and dissimilar aircraft pairs}
	\label{fig:SimMatrices}
\end{figure}

Other than the three metric learning methods, another important element in the study is the identification of constraints. Specifically, for each environmental impact output $O$ (e.g., departure fuel burn), depending on the metric learning method, we may need to identify a set of similar aircraft $\mathcal{S}$, a set of dissimilar aircraft $\mathcal{D}$, and a set of relative constraints $\mathcal{R}$. In this study, MMC and ITML require the identification of sets $\mathcal{S}$ and $\mathcal{D}$. For this part we propose a new approach to identify sets $\mathcal{S}$ and $\mathcal{D}$ within the representative aircraft set (scope 3 in Figure~\ref{fig:Scopes}). We denote the set of representative aircraft types as $\Omega = \{\boldsymbol{A}_{1},...,\boldsymbol{A}_{N}\}$. This statistical identification process is depicted in Figure~\ref{fig:SimMatrices}. Here, we first define three matrices that are used in the identification process:
\begin{itemize}
    \item \textbf{Matrix $D$}: describes the pairwise Euclidean distances, computed using aircraft input features, between all pairs of aircraft types in $\Omega$. Its entries are standardized such that they have zero mean and unit variance.
    \item \textbf{Matrix $E$}: describes the pairwise (absolute) environmental impact differences between all pairs of aircraft types in $\Omega$. Its entries are standardized such that they have zero mean and unit variance. 
    \item \textbf{Matrix $F$}: describes the difference between the above two matrices. It is obtained through subtracting matrix $D$ from $E$, i.e., $F = E-D$ by subtracting the corresponding entries in matrices $E$ and $D$.
\end{itemize}

Figure~\ref{fig:SimMatrices} shows examples of matrices $D$, $E$, and $F$. The histogram below each matrix displays the distribution of its entries. Now we focus on the interpretation of matrix $F$, which is key to the identification of the set of similar aircraft $\mathcal{S}$ and the set of dissimilar aircraft $\mathcal{D}$. Because the values in both matrices $D$ and $E$ have been standardized to have zero mean and unit variance, entries in matrix $F$ measure the differences between ``Euclidean relative distances'' and ``actual relative distances''. For example, for a specific entry in matrix $F$, $f_{ij}$ is obtained through $f_{ij} = e_{ij} - d_{ij}$. If $f_{ij}$ is small and well below zero, it indicates that between aircraft $i$ and $j$, they have relatively small difference in this environmental impact yet relatively large Euclidean distance. In this case, aircraft $i$ and $j$ are supposed to have similar environmental impact, yet are deemed dissimilar from the Euclidean distance's perspective. In metric learning, such aircraft pairs should be enforced as similar aircraft pairs to correct the bias induced by Euclidean distance. On the other hand, if $f_{ij}$ is large and well above zero, it indicates that aircraft $i$ and $j$ are highly dissimilar in the environmental impact yet their Euclidean distance is too small to reflect such difference. Then, such aircraft pairs should be enforced as dissimilar aircraft pairs in metric learning. The bottom right plot in Figure~\ref{fig:SimMatrices} shows the histogram of entries in matrix $F$, where the negative and positive values generally indicate similar and dissimilar aircraft pairs, respectively. Here, instead of using all negative/positive entries in matrix $F$ as similar/dissimilar aircraft pairs, we only utilize the most recognizable top and bottom 10\% as constraints in metric learning.

Let $f_{(10)}$ and $f_{(90)}$ be the 10-th and 90-th percentile of unique entries in matrix $F$, we formally define the set of similar aircraft $\mathcal{S}$ and the set of dissimilar aircraft $\mathcal{D}$ as follows:
\begin{equation}
    \mathcal{S} = \left\{ \left(\boldsymbol{A}_{i}, \boldsymbol{A}_{j}\right) ~|~  \boldsymbol{A}_{i}, \boldsymbol{A}_{j} \in \Omega, ~ f_{ij} \leq f_{(10)}, ~ f_{ij} < 0\right\}
\end{equation}
\begin{equation}
    \mathcal{D} = \left\{ \left(\boldsymbol{A}_{i}, \boldsymbol{A}_{j}\right) ~|~  \boldsymbol{A}_{i}, \boldsymbol{A}_{j} \in \Omega, ~ f_{ij} \geq f_{(90)}, ~ f_{ij} > 0\right\}
\end{equation}

The identification of the set of relative constraints $\mathcal{R}$ can be enabled by data reduction. For a certain environmental impact output $O$, when all $N$ aircraft in $\Omega$ are ordered in ascending order, we obtain a sequence similar to order statistics $\mathcal{O} = \{O_{(1)} \leq O_{(2)} \leq ... \leq O_{(N)}\}$. Let $\boldsymbol{A}_{(i)}$ be the aircraft (feature vector) which corresponds to environmental impact output $O_{(i)}$, and the mapping between them $O_{(i)} = \mathbf{M} (\boldsymbol{A}_{(i)})$ and $\boldsymbol{A}_{(i)} = \mathbf{M}^{-1} (O_{(i)})$, we arrive at a set of aircraft ordered by the environmental impact output, $\mathcal{A} = \{\boldsymbol{A}_{(1)}, \boldsymbol{A}_{(2)}, ..., \boldsymbol{A}_{(N)}\}$. In a data reduction process, we select a subset of aircraft types while maintaining the same data distribution in $O$~\citep{gao2022prem}, which can best maintain the statistical integrity of the population. Define a reduction ratio $\rho$, this distributional population reduction process selects one from every $\rho$ aircraft in sequence $\mathcal{A}$. A reduced set of aircraft is therefore given by
\begin{equation}
    \Omega_\rho = \left\{\boldsymbol{A}_{(i)} ~|~ \boldsymbol{A}_{(i)} \in \mathcal{A}, ~ (i~~\text{mod}~~\rho) = 0, ~ 1 \leq i \leq N\right\}
\end{equation}

Here we want to consider the relative constraints at both microscopic (MI) and macroscopic (MA) levels. Therefore, we include aircraft triplets at two different levels of granularity: $\rho_{MI}$ and $\rho_{MA}$. The values of $\rho_{MI}$ and $\rho_{MA}$ depend on the size of $\Omega$. In our case study, $\rho_{MI} = 2$ and $\rho_{MA} = 5$ are appropriate levels. For each level of $\rho$, the set of aircraft triplets includes both forward (``+'') and backward (``--'') aircraft sequences. The set of relative constraints $\mathcal{R}$ is given by
\begin{equation}
    \mathcal{R} = \mathcal{R}_{MI, +} \cup \mathcal{R}_{MI, -} \cup \mathcal{R}_{MA, +} \cup \mathcal{R}_{MA, -}
\end{equation}
where
\begin{equation}
    \begin{aligned}
    &\mathcal{R}_{MI, +} = \left\{\left(\boldsymbol{A}_{(i)}, \boldsymbol{A}_{(i+1)}, \boldsymbol{A}_{(i+2)}\right) ~|~ \boldsymbol{A}_{(i)}, \boldsymbol{A}_{(i+1)}, \boldsymbol{A}_{(i+2)} \in \Omega_{\rho_{MI}}\right\}\\
    &\mathcal{R}_{MI, -} = \left\{\left(\boldsymbol{A}_{(i+2)}, \boldsymbol{A}_{(i+1)}, \boldsymbol{A}_{(i)}\right) ~|~ \boldsymbol{A}_{(i+2)}, \boldsymbol{A}_{(i+1)}, \boldsymbol{A}_{(i)} \in \Omega_{\rho_{MI}}\right\}\\
    &\mathcal{R}_{MA, +} = \left\{\left(\boldsymbol{A}_{(i)}, \boldsymbol{A}_{(i+1)}, \boldsymbol{A}_{(i+2)}\right) ~|~ \boldsymbol{A}_{(i)}, \boldsymbol{A}_{(i+1)}, \boldsymbol{A}_{(i+2)} \in \Omega_{\rho_{MA}}\right\}\\
    &\mathcal{R}_{MA, -} = \left\{\left(\boldsymbol{A}_{(i+2)}, \boldsymbol{A}_{(i+1)}, \boldsymbol{A}_{(i)}\right) ~|~ \boldsymbol{A}_{(i+2)}, \boldsymbol{A}_{(i+1)}, \boldsymbol{A}_{(i)} \in \Omega_{\rho_{MA}}\right\}
    \end{aligned}
\end{equation}

This concludes the descriptions of metric learning key concepts, three selected algorithms, and identification of constraints. We next apply these techniques in a complete case study on aircraft environmental impact segmentation.

\section{The Case Study on Aircraft Environmental Impacts}\label{sec:casestudy} 

Aircraft segmentation has recently emerged as a challenge in aviation environmental impact analysis. As introduced in Section~\ref{sec:litrev}, among the thousands of unique aircraft types operated around the world, people can only build detailed APM for a small amount of aircraft types for environmental impact modeling. An effective aircraft segmentation process can contribute to: (1) the selection of a subset representative aircraft types to sufficiently cover the richness and complexity of the population, and (2) the identification of ``proxy aircraft'' for those aircraft types without APM. 

The process of aircraft segmentation is fundamentally the same as unsupervised learning method clustering, which partitions all aircraft types into clusters such that aircraft types within a cluster are highly similar to one another and dissimilar to aircraft types in other clusters. The most crucial thing is that, the clustering process is supposed to group together aircraft types that have similar environmental impact characteristics, instead of those that are just similar in the available aircraft features. The choice of distance metric plays a vital role in this process. Euclidean distance and the more general Minkowski distance treat each aircraft feature equally in the computation of aircraft dissimilarity. Since some aircraft features are more important than others and should be assigned with higher significance, these conventional distance metrics may not accurately reflect the actual aircraft dissimilarity in environmental impact. In this case study, we use metric learning in a weakly-supervised manner to infuse some partial information from environmental impact outputs into the Mahalanobis distance. By using the learned Mahalanobis distance metrics in clustering, the aim is to group together aircraft types that are more homogeneous in environmental impact. 

Another important aspect to investigate is the generalization property of the learned Mahalanobis distance metrics. We hope that a Mahalanobis distance metric learned from a subset of aircraft types can be generalized and applied to the entire population. For example, in this study we include 214 aircraft types that are within the ANP database because only they have the ``ground truth'' in environmental impact outputs. We train the Mahalanobis distance metrics using a subset of this population and hope that they apply well to all 214 aircraft types. In this study, we select a subset with 86 aircraft types (40\%) for training the tailored Mahalanobis distance metrics. This proportion is selected because of two reasons. First, there are currently close to 300 aircraft types in the ANP database, while the number of aircraft types in need of frequent environmental impact modeling is estimated to be less than a thousand. 40\% is a valid ratio to start with in this generalization (or extrapolation) experiment. Second, this generalization is more difficult at the initial stage, i.e., learning tailored distance metrics from 86 aircraft types and generalizing to 214 aircraft types is more challenging than learning tailored distance metrics from 400 aircraft types and generalizing to 1,000 aircraft types. A good result at this level is a strong indicator of the approach's potential on the real-world scope. In addition, in the identification of similar and dissimilar aircraft sets, we only use a small amount of information as described in Section~\ref{sec:method}. This is to test whether the metric learning approach can be used in a weakly-supervised manner on this specific problem.

For each evaluation in this section, we compare results from the following five methods, including two baseline methods and three metric learning methods:
\begin{itemize}
    \item \textbf{Baseline 1}: Euclidean distance
    \item \textbf{Baseline 2}: Mahalanobis distance with covariance matrix
    \item \textbf{Metric learning method 1}: MMC, with the proposed statistical constraint sets
    \item \textbf{Metric learning method 2}: ITML, with the proposed statistical constraint sets
    \item \textbf{Metric learning method 3}: LMNN
\end{itemize}

On the implementation of metric learning algorithms, the source codes of many mature metric learning algorithms have been published online by the authors~\citep{metric-learn}. We utilize the publicly available implementations together with the proposed statistical constraints sets, apply them in the designed case study, and compare the results. In the following subsections, we apply the two baseline methods and the three metric learning methods on aircraft segmentation for the 6 environmental impact outputs. The performances of the five methods are then evaluated and compared using different means, including data visualization in Section~\ref{sec:casestudy1} and quantitative measures in Section~\ref{sec:casestudy2}. At the end of this section, Section~\ref{sec:casestudy3} discusses how the feature selection results from the learned metrics can contribute to some additional insights for the aircraft environmental impact segmentation problem.

\subsection{Visual Intuitions}\label{sec:casestudy1}

The visual comparison part mainly includes dimensionality reduction results via t-Distributed Stochastic Neighbor Embedding (t-SNE)~\citep{van2008visualizing}. Due to the space constraint, we cannot display the complete visual results for all environmental impact outputs. We use the result from the first environmental impact output in the list -- departure fuel burn (D\_Fuel) as a demonstration. Next in Section~\ref{sec:casestudy2} and Section~\ref{sec:casestudy3}, we will present the quantitative evaluation results for all environmental impact outputs.

\begin{figure}[h!]
     \centering
     \begin{subfigure}[b]{\textwidth}
         \centering
         \includegraphics[width=0.4\textwidth]{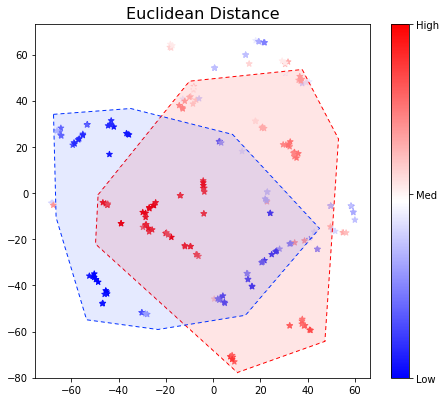}
         \caption{t-SNE visualization of aircraft representations: Euclidean distance}
         \label{fig:tSNE1}
     \end{subfigure}\vspace*{0.2cm}
     \begin{subfigure}[b]{\textwidth}
         \centering
         \includegraphics[width=0.4\textwidth]{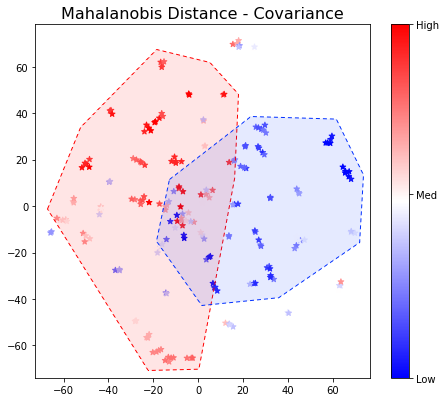}
         \hspace*{0.3cm}
         \includegraphics[width=0.4\textwidth]{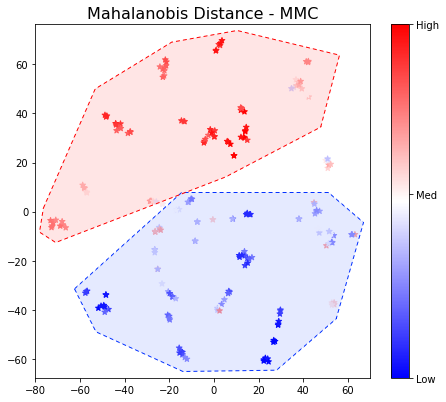}\\
         \vspace*{0.2cm}
         \includegraphics[width=0.4\textwidth]{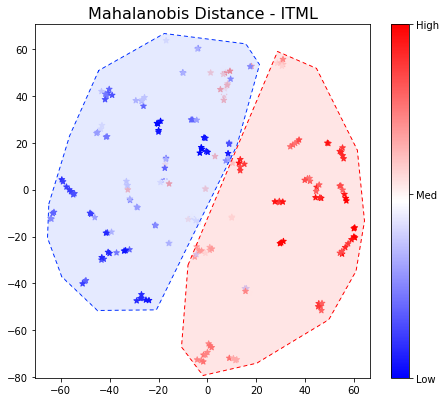}
         \hspace*{0.3cm}
         \includegraphics[width=0.4\textwidth]{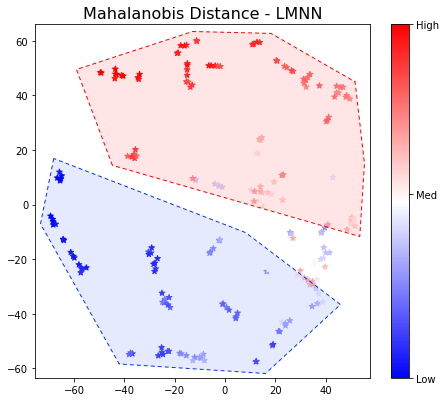}
         \caption{t-SNE visualization of aircraft representations: Mahalanobis distance with covariance matrix, MMC, ITML, and LMNN}
         \label{fig:tSNE2}
     \end{subfigure}
     \caption{Comparison of t-SNE visualizations for departure fuel burn}
     \label{fig:tSNEs}
\end{figure}

Figure~\ref{fig:tSNEs} displays the t-SNE visualizations of the complete aircraft population under the five different distance metrics. For MMC, ITML, and LMNN, the corresponding Mahalanobis distance metrics are learned from partial information in the training set. t-SNE is a nonlinear dimensionality reduction technique and one of the most recognized techniques for embedding and visualizing high-dimensional data in a low-dimensional space. In our study, each aircraft type is a 66-dimensional data point (the number of aircraft features). Here, t-SNE visualizes this high-dimensional data via giving each data point a location in the two-dimensional map while revealing the structure of the population at various scales. Taking the pairwise distance function as input, t-SNE first constructs a joint probability distribution $P$ on the aircraft's pairwise similarities in the high-dimensional space and uses Student's t-distribution to compute a similar distribution $Q$ over the aircraft's counterparts in the low-dimensional space. Then it minimizes the Kullback-Leibler (KL) divergence between $P$ and $Q$ with respect to the locations of aircraft types in the low-dimensional map. The low-dimensional map created by t-SNE captures the local structure of the high-dimensional data well, while also preserving some important global structure of the population such as the presence of clusters. As a result, similar aircraft types are expected to be modeled by nearby points in the low-dimensional map.

In Figure~\ref{fig:tSNEs}, each dot is an aircraft type, and we use color coding to denote the relative magnitude of its departure fuel burn, where blue indicates low and red indicates high. Figure~\ref{fig:tSNE1} first shows the t-SNE map of the 214 aircraft types under the Euclidean distance. Overall, because t-SNE visualization helps display the global structure of the population in a low-dimensional space, it is expected that a good distance metric should result in adequate separation between clusters with high and low departure fuel burns. In other words, the distinct red and blue dots should in general form two separated clusters under a good distance metric that can effectively reflect the aircraft population's departure fuel burn. In Figure~\ref{fig:tSNE1}, however, we observe that Euclidean distance fails to capture the pattern of departure fuel burn in the population, as the red and blue dots are mixed together without forming distinct clusters. Next, we look at the Mahalanobis distance results in Figure~\ref{fig:tSNE2}. In the upper left plot of Figure~\ref{fig:tSNE2}, we can see that under the unsupervised Mahalanobis distance with covariance matrix, the red and blue dots form their own clusters to certain degrees, yet the red and blue clusters still have an overlap in the central region. The rest of three plots in Figure~\ref{fig:tSNE2} display results from the three supervised metric learning approaches -- MMC, ITML, and LMNN. In these three results, the clusters with red and blue dots are generally well separated, except for some light-colored dots that have medium departure fuel burn levels. Results in Figure~\ref{fig:tSNEs} indicate that the three weakly-supervised Mahalanobis distance metrics are able to better capture the aircraft population's characteristics on departure fuel burn, which is desired for aircraft fuel burn segmentation. Before the use of quantitative measures in Section~\ref{sec:casestudy2}, these visual comparisons reveal how the tailored distance metrics affect the distribution of aircraft types and the proximity between them.

\subsection{Quantitative Measures}\label{sec:casestudy2}

In this subsection, we conduct an aircraft segmentation experiment and apply quantitative measures to assess the results. Through clear quantitative evidences, we aim to answer ``whether the tailored distance metrics can help identify more cohesive aircraft clusters for environmental impact outputs''. We apply the five distance metrics to partition the complete set of 214 aircraft types into different numbers of clusters. At each number of cluster $k$, we compare the intra-cluster aircraft homogeneity on the 6 environmental impact outputs. Hierarchical clustering (with average linkage) is employed in this process because of its deterministic results and recognition in many scientific applications, such as~\citep{costa2019adaptive}. The following two quantitative measures are used to assess the intra-cluster aircraft homogeneity on the environmental impact outputs:
\begin{itemize}
    \item \textbf{Coefficient of Variation (CV)}. The coefficient of variation is a statistical measure of dispersion. Within each cluster, the CV of an environmental impact output is the ratio of the standard deviation ($\sigma$) to the mean ($\mu$). It compares two clustering results and indicates `overall' which clustering has better intra-cluster homogeneity on an environmental impact output. At each number of cluster $k$, we will use the distribution of $k$ CVs in the assessment.
    \item \textbf{Maximum Range (MR)}. The range of an environmental impact output within a cluster is the difference between its largest and smallest values. Then, MR is the maximum among all $k$ cluster ranges. Compared to CV which is dimensionless, MR is a measure of the `worst case dispersion' and follows a non-increasing trend as $k$ increases. For both CV and MR, a smaller value indicates better intra-cluster homogeneity and therefore a better clustering result.
\end{itemize}

\begin{figure}[h!]
     \centering
     \begin{subfigure}[b]{\textwidth}
         \centering
         \includegraphics[width=0.24\textwidth]{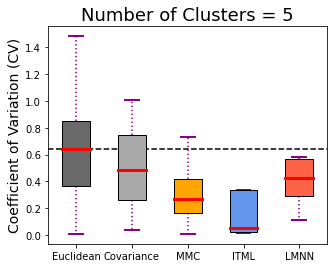}
         \includegraphics[width=0.24\textwidth]{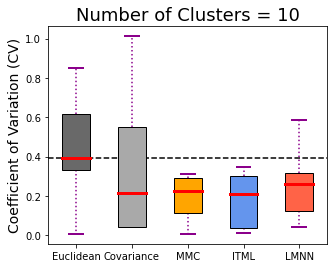}
         \includegraphics[width=0.24\textwidth]{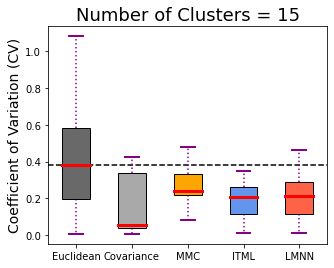}
         \includegraphics[width=0.24\textwidth]{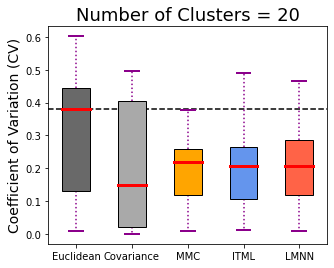}
         \caption{Departure fuel burn}
         \label{fig:CV_Fuel}
     \end{subfigure}\vspace*{0.3cm}
     \begin{subfigure}[b]{\textwidth}
         \centering
         \includegraphics[width=0.24\textwidth]{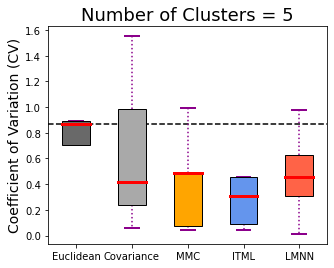}
         \includegraphics[width=0.24\textwidth]{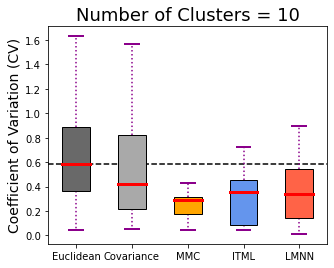}
         \includegraphics[width=0.24\textwidth]{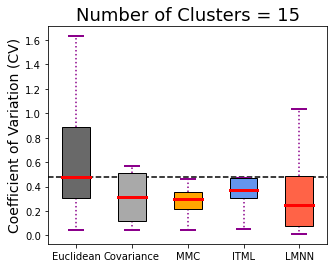}
         \includegraphics[width=0.24\textwidth]{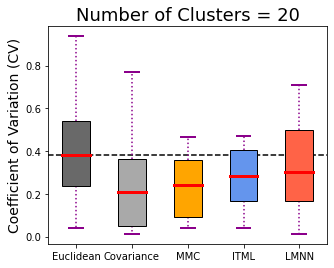}
         \caption{Departure NO$_x$ emissions}
         \label{fig:CV_NOX}
     \end{subfigure}\vspace*{0.3cm}
     \begin{subfigure}[b]{\textwidth}
         \centering
         \includegraphics[width=0.24\textwidth]{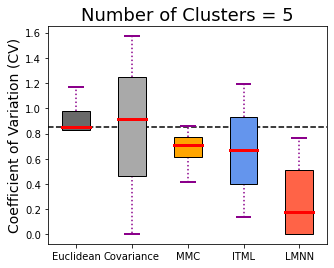}
         \includegraphics[width=0.24\textwidth]{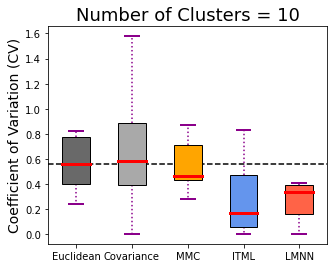}
         \includegraphics[width=0.24\textwidth]{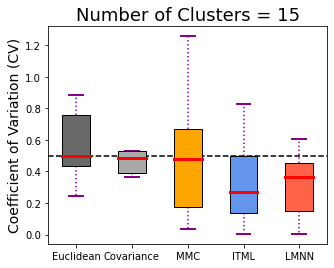}
         \includegraphics[width=0.24\textwidth]{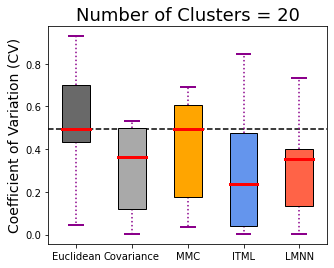}
         \caption{Departure CO emissions}
         \label{fig:CV_CO}
     \end{subfigure}\vspace*{0.3cm}
     \begin{subfigure}[b]{\textwidth}
         \centering
         \includegraphics[width=0.24\textwidth]{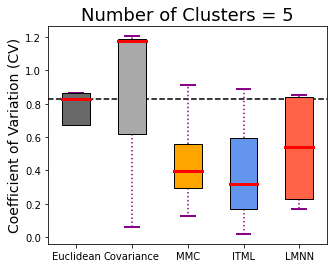}
         \includegraphics[width=0.24\textwidth]{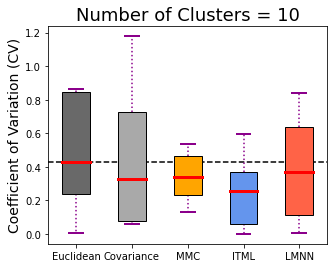}
         \includegraphics[width=0.24\textwidth]{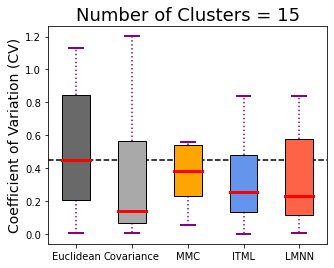}
         \includegraphics[width=0.24\textwidth]{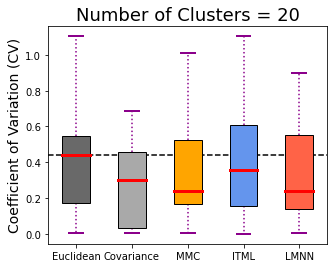}
         \caption{Departure nvPM emissions}
         \label{fig:CV_NVPM}
     \end{subfigure}
     \caption{Distribution of CV at different number of aircraft clusters: departure fuel burn and emissions}
     \label{fig:CV-FandE}
\end{figure}

Figure~\ref{fig:CV-FandE} shows comparisons of CV for fuel burn and emissions outputs. We use boxplot to visualize the distribution of CV for each clustering. Each boxplot displays the distribution's median, $Q_1$ (25-th percentile), $Q_3$ (75-th percentile), interquartile range ($IQR$), minimum ($Q_1 - 1.5*IQR$), and maximum ($Q_3 + 1.5*IQR$). In each plot in Figure~\ref{fig:CV-FandE}, results from the two baselines, Euclidean distance and Mahalanobis distance with covariance matrix, are colored in black and gray, respectively, on the left; results from the three metric learning methods, MMC, ITML, and LMNN, are colored in yellow, blue, and red, respectively, on the right. In addition, a horizontal dashed line depicts the median of the Euclidean distance result and serves as a reference level.

Figure~\ref{fig:CV_Fuel} first shows the result for departure fuel burn. It is obvious that the three metric learning methods have a clear edge over the Euclidean distance on departure fuel burn. At different numbers of clusters, the distributions of CV produced by MMC, ITML, and LMNN concentrate at smaller values -- their 75-th percentiles are much lower than the median of the distribution produced by the Euclidean distance. This indicates that the three tailored distance metrics learned from partial information can help identify aircraft types that are more similar in departure fuel burn. It is worth noticing that the unsupervised method, Mahalanobis distance with covariance matrix, also has an edge over the Euclidean distance on departure fuel burn. Figure~\ref{fig:CV_NOX} shows the result for departure NO$_x$ emissions, whose overall patterns are similar to the departure fuel burn result. The only exception is that when $k = 20$, we observe from the rightmost plot of Figure~\ref{fig:CV_NOX} that the advantage of LMNN is relatively small. Figure~\ref{fig:CV_CO} shows the result for departure CO emissions. On CO, we can observe that while ITML and LMNN still have clear advantages over the Euclidean distance, the advantage of MMC is relative small when $k$ becomes larger. In addition, Mahalanobis distance with covariance matrix is outperformed by the Euclidean distance when $k$ is small. Lastly, Figure~\ref{fig:CV_NVPM} shows the result for departure nvPM emissions. On nvPM, we can see that although the three metric learning methods are still better in general, the overall advantage is smaller than what are observed in fuel burn and NO$_x$. When $k$ increases to 20, although the metric learning methods still produce lower median values, the 25-th and 75-th percentiles are very close to the Euclidean distance result. 

\begin{figure}[h!]
     \centering
     \begin{subfigure}[b]{\textwidth}
         \centering
         \includegraphics[width=0.24\textwidth]{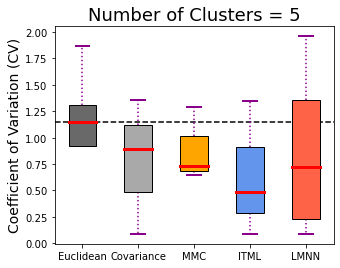}
         \includegraphics[width=0.24\textwidth]{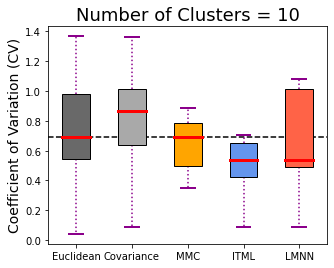}
         \includegraphics[width=0.24\textwidth]{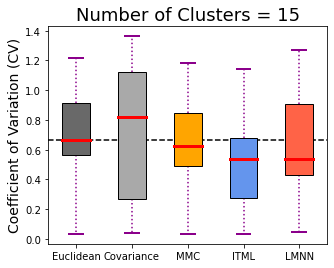}
         \includegraphics[width=0.24\textwidth]{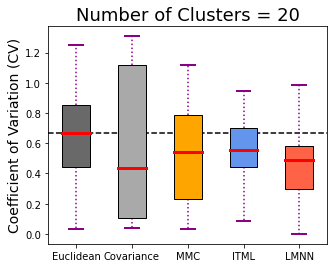}
         \caption{Departure SEL contour area at 80 dB}
         \label{fig:CV_N80}
     \end{subfigure}\vspace*{0.3cm}
     \begin{subfigure}[b]{\textwidth}
         \centering
         \includegraphics[width=0.24\textwidth]{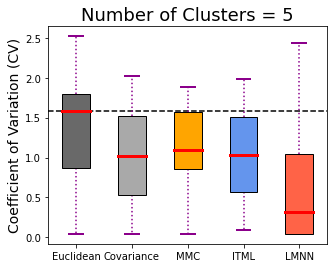}
         \includegraphics[width=0.24\textwidth]{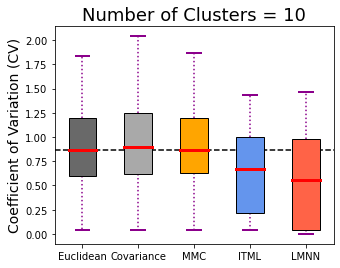}
         \includegraphics[width=0.24\textwidth]{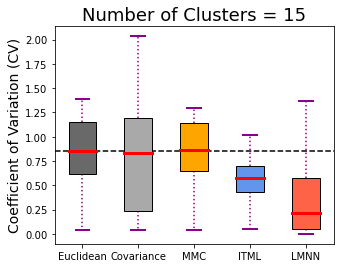}
         \includegraphics[width=0.24\textwidth]{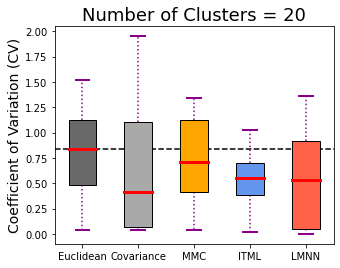}
         \caption{Departure SEL contour area at 90 dB}
         \label{fig:CV_N90}
     \end{subfigure}
     \caption{Distribution of CV at different number of aircraft clusters: departure noise contour areas}
     \label{fig:CV-N}
\end{figure}

We then look at the comparisons of CV for noise outputs in Figure~\ref{fig:CV-N}. Overall on noise outputs, due to the lack of noise-related aircraft features (with further discussions in Section~\ref{sec:remarks}), we can see that metric learning is less effective compared to the fuel burn and emissions results in Figure~\ref{fig:CV-FandE}. Among the three metric learning methods, ITML is one that can consistently perform well in Figure~\ref{fig:CV-N}. LMNN produces good result in 90 dB SEL contour area yet is inconsistent in 80 dB SEL contour area; MMC performs better in 80 dB SEL contour area yet has no clear advantage in 90 dB SEL contour area. 

\begin{figure}[h!]
	\centering
	\includegraphics[width=0.45\textwidth]{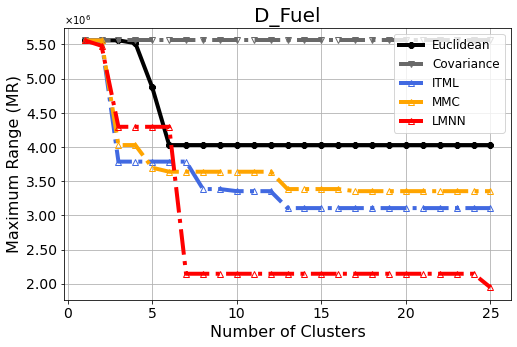}
        \hspace*{0.8cm}
        \includegraphics[width=0.45\textwidth]{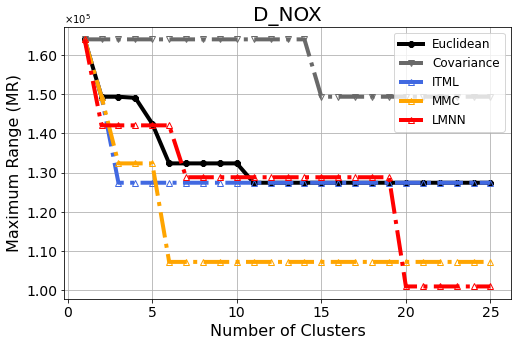}\\
        \vspace*{0.4cm}
        \includegraphics[width=0.45\textwidth]{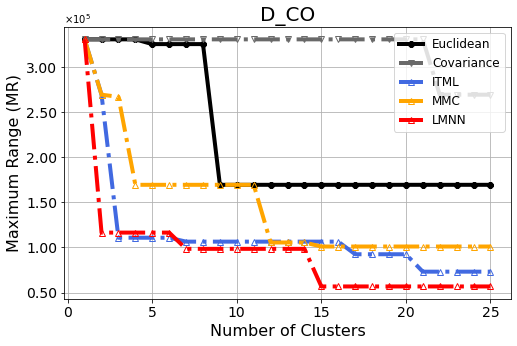}
        \hspace*{0.8cm}
        \includegraphics[width=0.45\textwidth]{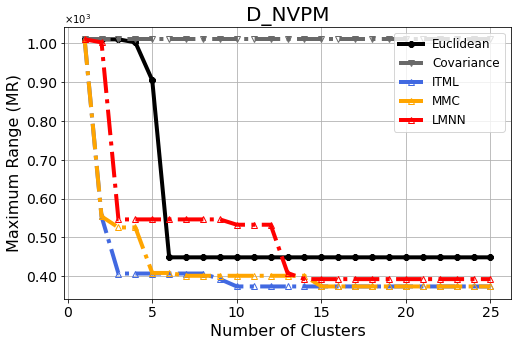}
	\caption{MR vs. number of aircraft clusters: departure fuel burn and emissions}
	\label{fig:MR-FandE}
\end{figure}

\begin{figure}[h!]
	\centering
	\includegraphics[width=0.45\textwidth]{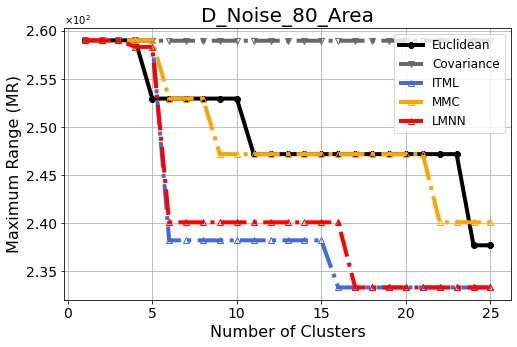}
        \hspace*{0.8cm}
        \includegraphics[width=0.45\textwidth]{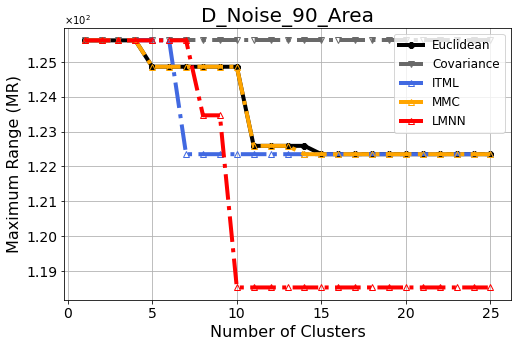}
	\caption{MR vs. number of aircraft clusters: departure noise contour areas}
	\label{fig:MR-N}
\end{figure}

The second quantitative measure MR provides another angle to evaluate the aircraft segmentation results. Under the same color coding, Figure~\ref{fig:MR-FandE} shows how MR changes with the number of aircraft clusters for departure fuel burn and emissions. The first observation here is that, although Mahalanobis distance with covariance matrix can provide some decent results in CV, it has the worst performance among all five distance metrics in MR. This reflects a limitation of unsupervised metric learning in this case study. In Figure~\ref{fig:MR-FandE}, the Euclidean distance is dominated by metric learning methods MMC and ITML, because the yellow and blue curves are always at or under the black baseline curves. This is a strong indicator of their advantages in this measure. The performance of LMNN is strong in fuel burn and CO emissions. On NO$_x$ and nvPM emissions, the LMNN curve is above the Euclidean distance curve within some ranges of $k$. Figure~\ref{fig:MR-N} shows the MR results for departure noise outputs. LMNN performs strong in both noise outputs, although it does not dominate the Euclidean distance at very few small $k$ values. ITML also has reliable performance on the two noise outputs and even dominates LMNN in 80 dB SEL contour area. MMC has comparable performance with the Euclidean distance on the two noise outputs. 

Below is a summary of the main takeaways from this aircraft segmentation case study:
\begin{itemize}
    \item[(a)] Metric learning does contribute to improved performance in aircraft environmental segmentation. In most scenarios (environmental impact output + number of clusters), metric learning helps generate aircraft clusterings that have lower CV and MR values, indicating that the process can now group together aircraft types that have similar environmental impact characteristics.
    \item[(b)] The effectiveness of metric learning varies with different environmental impact outputs. In this case study, metric learning overall has a clear advantage on fuel burn, NO$_x$ emissions, and CO emissions, and a relatively small advantage on nvPM emissions and the two noise outputs. The performance on noise outputs is partly due to a lack of noise-related aircraft features on the input side.
    \item[(c)] The unsupervised method Mahalanobis distance with covariance matrix has major limitations. Its performance on CV can be comparable to those of the three supervised metric learning methods in some cases; in other cases, it is outperformed by the Euclidean distance. Its performance on MR is the worst among all five distance metrics. 
    \item[(d)] Among the three supervised metric learning methods, ITML is one that can consistently perform well in this case study. MMC and LMNN also have strong performances in some scenarios. Because ITML is a weakly-supervised metric learning method whose only parameter is easy to optimize, overall it may have superiority in the aircraft segmentation problem.
\end{itemize}

\subsection{Feature Selection}\label{sec:casestudy3}

In this subsection, we extract feature importance insights from the interpretable metric learning method MMC. In Section~\ref{sec:method}, we mentioned that MMC allows the restriction of matrix $M$ in Mahalanobis distance to be diagonal. Then, the Mahalanobis distance becomes a weighted Euclidean distance function that assigns different weights to different aircraft features. Because all aircraft features have been standardized to the same scale, a higher weight outputted by MMC indicates that the corresponding aircraft feature is more significant and influential during aircraft segmentation for the environmental impact output.

\begin{figure}[h!]
	\centering
	\includegraphics[width=0.47\textwidth]{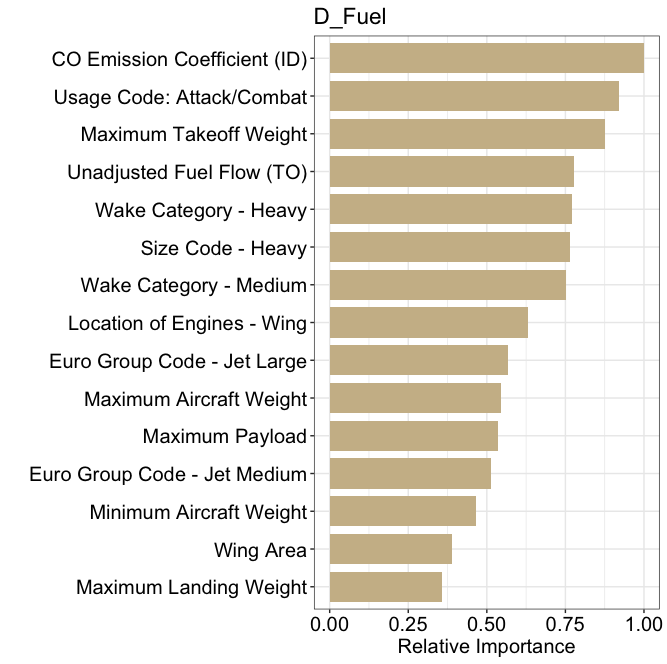}
        \hspace*{0.6cm}
        \includegraphics[width=0.47\textwidth]{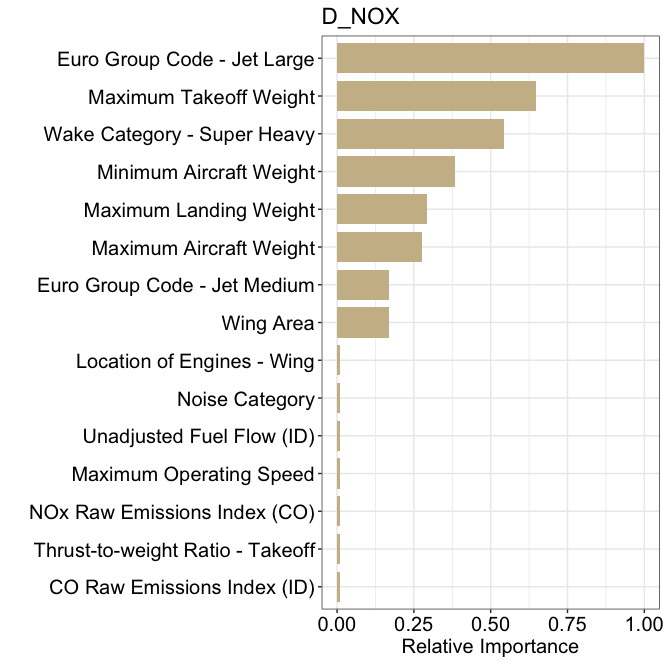}\\
        \vspace*{0.3cm}
        \includegraphics[width=0.47\textwidth]{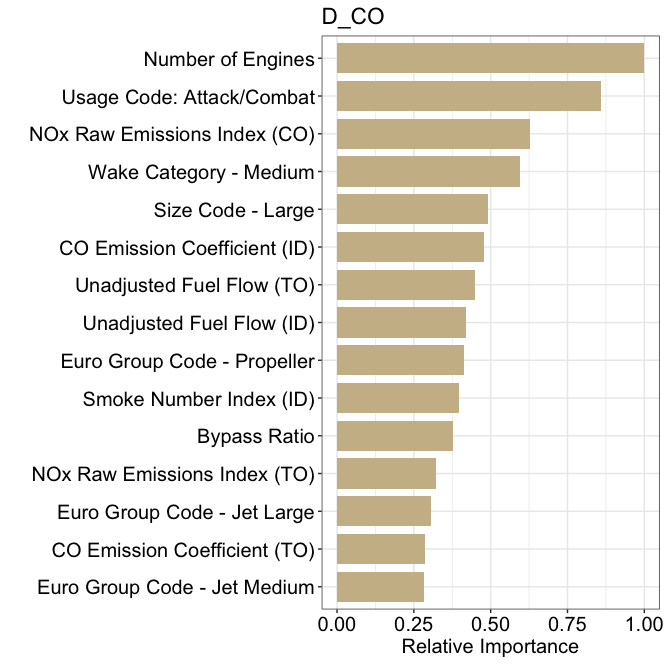}
        \hspace*{0.6cm}
        \includegraphics[width=0.47\textwidth]{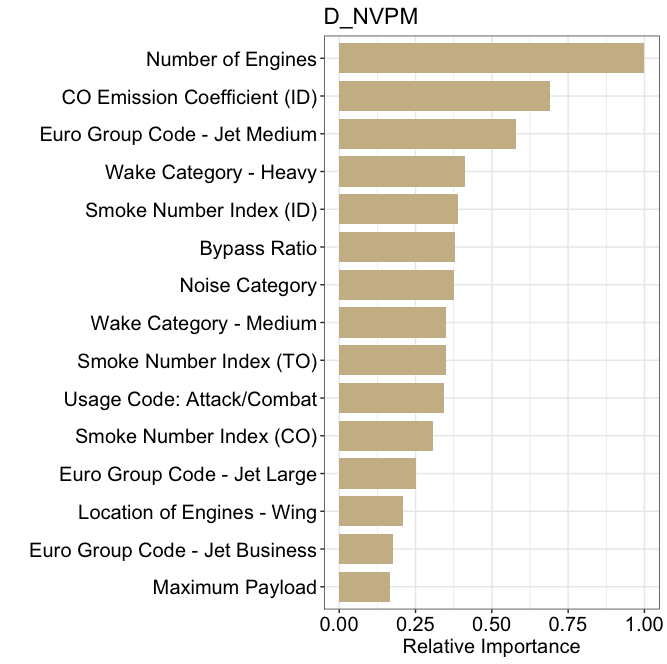}
	\caption{Top 15 features and their relative importance scores: departure fuel burn and emissions}
	\label{fig:FI-FandE}
\end{figure}


The MMC feature selection result often reflects sparsity, i.e., zero weight is assigned to a certain number of features that are `unused' in distance computation. For each environmental impact output, we present the top 15 features and their relative importance scores. Figure~\ref{fig:FI-FandE} displays the feature importance results for the four fuel burn and emissions outputs. The upper left plot of Figure~\ref{fig:FI-FandE} shows the top 15 important aircraft features for the segmentation of aircraft departure fuel burn. On this specific output, many aircraft features have close importance scores. Among the top 15 aircraft features for departure fuel burn, there are aircraft categorization features (e.g., Usage Code, Wake Category, Euro Group code), weight information, fuel and CO emission coefficients, and aircraft configuration parameters. In contrast, the departure NO$_x$ emissions result shown in the upper right plot of Figure~\ref{fig:FI-FandE} only highlights 8 features that are significantly more important than other aircraft features considered. These 8 aircraft features include aircraft categorizations (Euro Group code and Wake Category), weight information, and one aircraft configuration parameter (wing area). Therefore, aircraft size parameters are of vital significance in inferring an aircraft type's NO$_x$ emissions. In the CO emissions result shown in the lower left plot of Figure~\ref{fig:FI-FandE}, a larger proportion of the significant features are about aircraft engine, such as the engine number and bypass ratio, as well as fuel burn and emissions coefficients. Lastly, the lower right plot of Figure~\ref{fig:FI-FandE} shows a mixture of aircraft categorizations and engine features that are crucial for the segmentation of aircraft nvPM emissions. 

\begin{figure}[h!]
	\centering
	\includegraphics[width=0.47\textwidth]{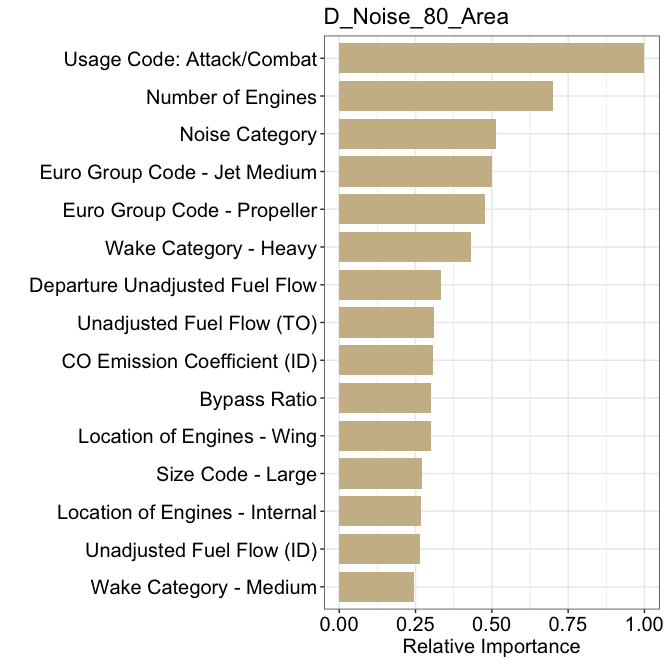}
        \hspace*{0.6cm}
        \includegraphics[width=0.47\textwidth]{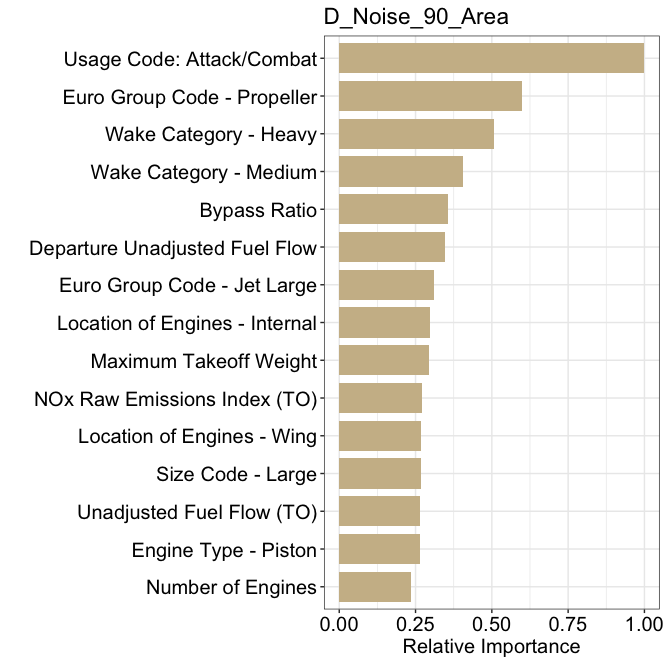}
	\caption{Top 15 features and their relative importance scores: departure noise contour areas}
	\label{fig:FI-N}
\end{figure}

Figure~\ref{fig:FI-N} displays the feature importance results for the two noise outputs. The results are similar between 80 dB and 90 dB SEL contour areas, as 10 common features appear in the two top 15 feature sets. The first observation here is that, [Usage Code: Attack/Combat] is selected as the most influential aircraft feature in both noise outputs. This indicates that attack aircraft is in its own class regarding aircraft noise and should be identified first in aircraft segmentation for departure noise. Engine-related features, such as engine number, bypass ratio, and location are also significant for identifying aircraft with similar departure noise properties. The other two influential factors are aircraft wake category and fuel flow coefficients. Overall, a feature selection practice can help practitioners better understand the interrelations between aircraft features and aircraft environmental impact outputs and the priorities for collecting accurate information. Consequently, an interpretable metric learning method embraces particular superiority in similar problems.

Below is a summary of the main takeaways from the feature selection practice:
\begin{itemize}
    \item[(a)] Different environmental impact outputs have different significant aircraft features in the aircraft segmentation process. To obtain precise aircraft segmentation results, a tailored distance metric should be learned and used for each individual environmental impact output.
    \item[(b)] Aircraft weight information, configuration parameters, and categorizations are influential in the segmentation of aircraft departure fuel burn and NO$_x$ emissions; aircraft engine features are influential for CO emissions.
    \item[(c)] For noise-related outputs, the first step in the aircraft segmentation process should identify whether an aircraft belongs to attack aircraft. After that, aircraft engine features and fuel flow coefficients can be used for more detailed segmentations.
\end{itemize}

\section{Remarks}\label{sec:remarks} 

This work is the very first study that uses an metric learning approach to improve the performance of aircraft segmentation for environmental impact. Despite a challenging case study due to: (1) the range and variability of aircraft types involved, (2) a small training set, and (3) the complexity in aviation environmental impact analysis, we still obtained many positive outcomes which demonstrate the potential of metric learning in related problems. Overall, we observed that by utilizing partial information, metric learning can indeed help aircraft segmentation better identify aircraft clusters that are similar in environmental impact outputs. However, it is also important to note that the present work is not a complete and final study on this topic. Continuous efforts are needed to further expand and advance the current progress. In this section we discuss some limitations and extensions of this work.

\subsection{Limitations}

The first limitation in this study lies in the range of aircraft features. Although the data collection process aims to include as many aircraft features as possible and is more comprehensive than relevant studies in the past, the list of features is still not close to being exhaustive. The approach's performance on noise outputs can be further improved through collecting additional features related to noise. One example data source is the noise certification data, which describes an aircraft type's noise level from multiple dimensions. In addition, the current feature set does not include parameters from the aircraft/airline operations side, such as departure and arrival procedures. The procedure information is usually in different formats and is relatively indirect to use, yet could become a great addition to the range of features if methodical approaches are employed to quantify and parameterize them. However, it must be noted that for features such as noise certification data, it could be difficult to collect them for a wide range of aircraft types. If an aircraft feature is only available for a small fraction of aircraft types, it would be of less practical use in relevant studies. A second limitation surrounds the global distance metric used in this study. Our current target is to, for each environmental impact output, learn a global Mahalanobis distance metric for all aircraft types. If there exists some natural clusters in the aircraft population, where in each cluster a subset of aircraft types are highly similar to each other on one or more environmental impacts, it could be beneficial to first divide the population and learn multiple local distance metrics. A ``hierarchical segmentation'' could potentially reinforce the advantage of metric learning and the experiment result even further. This point will also be further discussed next in the extensions. Lastly, the case study selects a representative setting on factors such as airport, profile, and weather. When using partial information to train tailored distance metrics, the choices of similar and dissimilar aircraft types can be improved through the consideration of uncertainties in these factors. This requires a comprehensive and independent study that involves experimental design, very large-scale experiments, and statistical analysis.

\subsection{Extensions}

We propose three future avenues of this study that could yield fruitful results in the near future. First, local metric learning should be further explored in a similar setting. All baseline and metric learning methods create a global distance metric which covers all aircraft types in the population. Although this ``ambitious'' plan results in overall positive performance, the aircraft population is heterogeneous and a single global metric may not be the best solution for capturing the complexity of the task. It could potentially be beneficial use multiple local metrics that vary across the population space. In that case, the aircraft population is first partitioned into some natural groups/classes (large twin aisle, regional jet, general aviation, military, etc.), and one distance metric is learned for each class. In the meantime, local metric learning also has its own challenges~\citep{bellet2013metric}, such as overfitting, consistency, and the pre-classification step. The second extension is multi-task learning. The objective of multi-task learning is to learn the shared information/model across multiple related tasks~\citep{gao2022multi}. Given a set of related tasks/outputs, multi-task metric learning can be employed to either (1) learn a metric for each task in a coupled fashion, thus improve the performance on all tasks, or (2) learn a shared metric for all tasks. In aviation environmental impact modeling, there is a potential need for multi-task metric learning. For example, when identifying a proxy aircraft, one might want to find a substitution model that is close on all noise-related outputs. Then, it will be beneficial to learn a common underlying distance metric for those environmental impacts simultaneously. Third, and the most challenging extension, is to answer the question of ``how much partial information is needed to achieve certain performance level''. In real-world applications, the side information can be collected from limited ground truth labels, human expert inputs, physics rules and domain knowledge, or a combination of these sources. In this study we use predefined values on the size of training set (40\%) and the proportion of aircraft pairs (top and bottom 10\%), yet the sensitivity of both settings can be further explored in a future work.

\section{Conclusion}\label{sec:conclusion} 

This paper presented a metric learning approach for tackling a recent challenge in sustainable aviation -- aircraft environmental impact segmentation. Compared to the old, unsupervised aircraft segmentation approach, we added a metric learning solution which makes better use of the partial aviation environmental impact information to more effectively group together aircraft types that have similar environmental impact. The proposed approach consists of representative aircraft selection, identification of constraints, and metric learning algorithms. In a comprehensive case study, we integrated aircraft features for a population of 214 aircraft types within the ANP database and run computer simulations to obtain their key environmental impact outputs. The aircraft segmentation experiment is conducted using two baseline distance metrics (Euclidean distance, Mahalanobis distance with covariance matrix) and three metric learning methods (MMC, ITML, LMNN). We used selective partial information from the representative aircraft types' environmental impact outputs to train tailored Mahalanobis distance metrics, and tested whether they can generalize to the segmentation of the entire aircraft population. Through quantitative measures CV and MR, we observed that in most scenarios, metric learning helps generate aircraft clusterings that have lower CV and MR values. This indicates that metric learning can help group together aircraft types that are more homogeneous in environmental impact characteristics. ITML is the method that can consistently perform well in the case study. A feature selection practice through MMC generated useful insights into the problem and highlighted important aircraft features for each environmental impact output. Going forward, this study can further include more aircraft features and consider more complex formulations such as local metric learning approaches, nonlinear metrics, and multi-task learning. The metric learning solution can also be applied to re-visit and refine a number of data-driven analytical studies in aviation and transportation. 

\section*{Acknowledgement}\label{sec:acknowledgement} 

The authors would like to thank Dr. Ameya Behere for his help on running AEDT experiment which generates data for this study. The authors would also like to thank Dr. Yao Xie for the helpful discussions and inspirations on metric learning which lead to this work. 

\newpage

\section*{Appendix: Details of the Aircraft Features, Environmental Impact Metrics, and the Range of Aircraft Types}

\begin{table}[h!]
\centering
\caption{The complete list of existing aircraft features (attribute type: Nu - numeric, No - nominal, Or - ordinal; stage code: TO - takeoff, CO - climb out, AP - approach, ID - idle)}
\label{tbl:ori-feature}
\begin{tabular}{cccccc}
\hline
\textbf{No.} & \textbf{Category}                         & \textbf{Code} & \textbf{Feature Name}      & \textbf{Type} & \textbf{Unit} \\ \hline
1               & \multirow{8}{*}{Airframe}                 & ENGINE\_COUNT           & Number of Engines            & Nu            & -             \\   
2               &                                           & ENGINE\_LOCATION        & Location of Engines          & No            & -             \\  
3               &                                           & DESIGNATION\_CODE       & Designation Code             & No            & -             \\  
4               &                                           & EURO\_GROUP\_CODE       & Euro Group Code              & No            & -             \\  
5               &                                           & MAX\_RANGE              & Maximum Range                & Nu            & n.mi.         \\  
6               &                                           & USAGE\_CODE             & Usage Code                   & No            & -             \\  
7               &                                           & SIZE\_CODE              & Size Code                    & Nu            & -             \\  
8               &                                           & ENGINE\_TYPE            & Engine Type                  & No            & -             \\ \hline
9               & \multirow{2}{*}{Engine}                   & BYPASS\_RATIO           & Bypass Ratio                 & Nu            & -             \\  
10              &                                           & PRESSURE\_RATIO         & Pressure Ratio               & Nu            & -             \\ \hline
11              & \multirow{13}{*}{Aircraft}                & NOISE\_CAT              & Noise Category               & Or            & -             \\  
12              &                                           & MX\_GW\_TKO             & Maximum Takeoff Weight       & Nu            & lbs           \\  
13              &                                           & MX\_GW\_LND             & Maximum Landing Weight       & Nu            & lbs           \\  
14              &                                           & COEFF\_TYPE             & Thrust Coefficient Type      & No            & -             \\  
15              &                                           & THR\_STATIC             & Static Thrust                & Nu            & lbs           \\  
16              &                                           & WAKE\_CAT\_CODE         & Wake Category                & Or            & -             \\  
17              &                                           & MASS\_REF               & Reference Aircraft Weight    & Nu            & lbs           \\  
18              &                                           & MASS\_MIN               & Minimum Aircraft Weight      & Nu            & lbs           \\  
19              &                                           & MASS\_MAX               & Maximum Aircraft Weight      & Nu            & lbs           \\  
20              &                                           & MASS\_PAYLD             & Maximum Payload              & Nu            & lbs           \\  
21              &                                           & FENV\_VMO               & Maximum Operating Speed      & Nu            & kts           \\  
22              &                                           & FENV\_ALT               & Maximum Operating Altitude   & Nu            & ft MSL        \\  
23              &                                           & WING\_AREA              & Wing Area                    & Nu            & $m^2$    \\ \hline
24              & \multirow{12}{*}{\begin{tabular}[c]{@{}c@{}}Fuel Burn\\ and Emissions\end{tabular}} & UA\_RWF\_TO             & Unadjusted Fuel Flow (TO)    & Nu            & kg/s          \\  
25              &                                           & UA\_RWF\_CO             & Unadjusted Fuel Flow (CO)    & Nu            & kg/s          \\  
26              &                                           & UA\_RWF\_ID             & Unadjusted Fuel Flow (ID)    & Nu            & kg/s          \\  
27              &                                           & CO\_REI\_TO             & CO Raw Emissions Index (TO)  & Nu            & g/kg          \\  
28              &                                           & CO\_REI\_CO             & CO Raw Emissions Index (CO)  & Nu            & g/kg          \\  
29              &                                           & CO\_REI\_ID             & CO Raw Emissions Index (ID)  & Nu            & g/kg          \\  
30              &                                           & NOX\_REI\_TO            & NOx Raw Emissions Index (TO) & Nu            & g/kg          \\  
31              &                                           & NOX\_REI\_CO            & NOx Raw Emissions Index (CO) & Nu            & g/kg          \\  
32              &                                           & NOX\_REI\_ID            & NOx Raw Emissions Index (ID) & Nu            & g/kg          \\
33              &                                           & SN\_TO                  & Smoke Number Index (TO)      & Nu            & -             \\  
34              &                                           & SN\_CO                  & Smoke Number Index (CO)      & Nu            & -             \\  
35              &                                           & SN\_ID                  & Smoke Number Index (ID)      & Nu            & -             \\  
36              &                                           & SN\_MAX                 & Maximum Smoke Number Index   & Nu            & -             \\ \hline
\end{tabular}
\end{table}

\begin{table}[htbp]
\centering
\caption{The complete list of aircraft features from feature engineering (all numeric)}
\label{tbl:eng-feature}
\begin{tabular}{ccccc}
\hline
\textbf{No.} & \textbf{Code}    & \textbf{Feature Name}                & \textbf{Formulation}                          & \textbf{Unit}  \\ \hline
1               & NOX\_COEFF\_TO   & NOx Emission Coefficient (TO)        & THR STATIC * NOX REI TO  & kg              \\ 
2               & NOX\_COEFF\_CO   & NOx Emission Coefficient (CO)        & THR STATIC * NOX REI CO  & kg              \\ 
3               & NOX\_COEFF\_ID   & NOx Emission Coefficient (ID)        & THR STATIC * NOX REI ID  & kg              \\ 
4               & CO\_COEFF\_TO    & CO Emission Coefficient (TO)         & THR STATIC * CO REI TO   & kg              \\ 
5               & CO\_COEFF\_CO    & CO Emission Coefficient (CO)         & THR STATIC * CO REI CO   & kg              \\ 
6               & CO\_COEFF\_ID    & CO Emission Coefficient (ID)         & THR STATIC * CO REI ID   & kg              \\ 
7              & TWR\_TO          & Thrust-to-weight Ratio - Takeoff & THR STATIC / MX GW TKO          & -              \\ 
8              & WING\_LOADING    & Wing Loading                         & MX GW TKO / WING AREA              & lbs/$m^2$ \\ 
9              & MLD\_MTOW\_RATIO & MLW-to-MTOW Ratio                    & MX GW LND / MX GW TKO & -              \\ 
10              & DEP\_FUEL\_FLOW  & Departure Unadjusted Fuel Flow       & ID - 75\%, TO - 5\%, CO - 20\%       & kg/s           \\ \hline
\end{tabular}
\end{table}

\begin{table}[htbp]
\centering
\caption{Levels and explanations for categorical features}
\label{tbl:cat-feature}
\begin{tabular}{cccc}
\hline
\textbf{No.} & \textbf{Code}     & \textbf{Feature Name}   & \textbf{Categorical Feature Description}                                                                                                                                           \\ \hline
1            & ENGINE\_LOCATION  & Location of Engines     & F - Fuselage/Tail, I - Internal, W - Wing                                                                                                                                          \\ \hline
2            & DESIGNATION\_CODE & Designation Code        & C - Civil, G - General Aviation, M - Military                                                                                                                                      \\ \hline
3            & EURO\_GROUP\_CODE & Euro Group Code         & \begin{tabular}[c]{@{}c@{}}JB - Jet Business, JL - Jet Large, JM - Jet Medium,\\ JR - Jet Regional, JS - Jet Small, PP - Propeller,\\ SS - Supersonic, TP - Turboprop\end{tabular} \\ \hline
4            & USAGE\_CODE       & Usage Code              & \begin{tabular}[c]{@{}c@{}}A - Attack/Combat, B - Business,\\ C - Cargo/Transport, O - Other, P - Passenger\end{tabular}                                                          \\ \hline
5            & SIZE\_CODE        & Size Code               & \begin{tabular}[c]{@{}c@{}}H - Heavy, L - Large, M - Medium,\\ S - Small, T - Light, V - Very Light\end{tabular}                                                                   \\ \hline
6            & ENGINE\_TYPE      & Engine Type             & \begin{tabular}[c]{@{}c@{}}E - Electric, J - Jet, P - Piston,\\ R - Rocket, T - Turboprop/Turboshaft\end{tabular}                                                                  \\ \hline
7            & NOISE\_CAT        & Noise Category          & Noise stage number - 0, 1, 2, 3, 4                                                                                                                                                 \\ \hline
8            & WAKE\_CAT\_CODE   & Wake Category           & \begin{tabular}[c]{@{}c@{}}J - Super Heavy, H - Heavy,\\ M -Medium, L - Light\end{tabular}                                                                                         \\ \hline
\end{tabular}
\end{table}

\begin{table}[htbp]
\centering
\caption{The complete list of all aircraft types used in the experiment}
\label{tbl:aircraft}
\begin{tabular}{ccccccccc}
\hline
1900D   & 747400     & A380-861    & CIT3     & DC870  & F105D    & LEAR25 \\
707     & 7478       & A4C         & CL600    & DC8QN  & F106     & LEAR35 \\
707120  & 747SP      & A6A         & CL601    & DC910  & F111AE   & MD11GE \\
707320  & 757300     & A7D         & CNA172   & DC930  & F117A    & MD11PW \\
707QN   & 757PW      & A7E         & CNA182   & DC93LW & F14A     & MD81   \\
717200  & 757RR      & ATR72-212A  & CNA206   & DC950  & F15E20   & MD82   \\
720B    & 767300     & B1          & CNA208   & DC95HW & F16GE    & MD83   \\
727100  & 767400     & B2A         & CNA20T   & DC9Q7  & F16PW0   & MD9025 \\
727200  & 767CF6     & BAC111      & CNA441   & DC9Q9  & F28MK2   & MD9028 \\
727D15  & 767JT9     & BAE146      & CNA500   & DHC6   & F28MK4   & MU3001 \\
727D17  & 777200     & BAE300      & CNA525C  & DHC6QP & F4C      & OV10A  \\
727EM1  & 777300     & BD-700-1A10 & CNA55B   & DHC7   & F5AB     & P3A    \\
727EM2  & 7773ER     & BD-700-1A11 & CNA560E  & DHC8   & F5E      & PA28   \\
727Q15  & 7878R      & BEC58P      & CNA560U  & DHC830 & FAL20    & PA30   \\
727Q7   & A10A       & C-130E      & CNA560XL & DO228  & FAL900EX & PA31   \\
727Q9   & A3         & C-20        & CNA680   & DO328  & FB111A   & PA42   \\
727QF   & A300-622R  & C118        & CNA750   & E3A    & G650ER   & S3A\&B \\
737     & A300B4-203 & C12         & COMJET   & EA6B   & GASEPF   & SABR80 \\
737300  & A310-304   & C130        & COMSEP   & EMB120 & GASEPV   & SD330  \\
737400  & A319-131   & C130AD      & CONCRD   & EMB145 & GII      & SF340  \\
737500  & A320-211   & C130E       & CRJ9-ER  & EMB14L & GIIB     & SR71   \\
737700  & A320-232   & C131B       & CRJ9-LR  & EMB170 & GIV      & T-2C   \\
737800  & A320-272N  & C135A       & CVR580   & EMB175 & GV       & T1     \\
737D17  & A321-232   & C135B       & DC1010   & EMB190 & HS748A   & T3     \\
737MAX8 & A330-301   & C141A       & DC1030   & EMB195 & IA1125   & T34    \\
737N17  & A330-343   & C17         & DC1040   & F-18   & KC135    & T42    \\
737N9   & A340-211   & C18A        & DC3      & F-4C   & KC135B   & U2     \\
74710Q  & A340-642   & C21A        & DC6      & F10062 & KC135R   & U21    \\
747200  & A350-941   & C23         & DC820    & F10065 & L1011    &        \\
74720A  & A37        & C5A         & DC850    & F100D  & L10115   &        \\
74720B  & A380-841   & C9A         & DC860    & F104G  & L188     &      \\ \hline
\end{tabular}
\end{table}

\newpage

\bibliography{Gao_Submission}

\begin{thebibliography}{63}
\newcommand{\enquote}[1]{``#1''}
\providecommand{\natexlab}[1]{#1}
\providecommand{\url}[1]{\texttt{#1}}
\providecommand{\urlprefix}{URL }
\expandafter\ifx\csname urlstyle\endcsname\relax
  \providecommand{\doi}[1]{doi:\discretionary{}{}{}#1}\else
  \providecommand{\doi}{doi:\discretionary{}{}{}\begingroup
  \urlstyle{rm}\Url}\fi

\bibitem[{Waitz et~al.(2004)Waitz, Townsend, Cutcher-Gershenfeld, Greitzer, and
  Kerrebrock}]{waitz2004aviation}
Waitz, I., Townsend, J., Cutcher-Gershenfeld, J., Greitzer, E., and Kerrebrock,
  J., \enquote{Aviation and the environment: A national vision statement,
  framework for goals and recommended actions,} 2004.

\bibitem[{{Federal Aviation Administration}(2015)}]{faa2015aviation}
{Federal Aviation Administration}, \enquote{Aviation emissions, impacts and
  mitigation: A primer,} \emph{Office of Environment and Energy}, 2015.

\bibitem[{Rosero et~al.(2007)Rosero, Ortega, Aldabas, and
  Romeral}]{rosero2007moving}
Rosero, J., Ortega, J., Aldabas, E., and Romeral, L., \enquote{Moving towards a
  more electric aircraft,} \emph{IEEE Aerospace and Electronic Systems
  Magazine}, Vol.~22, No.~3, 2007, pp. 3--9.

\bibitem[{Hughes et~al.(2011)Hughes, Van~Zante, and
  Heidmann}]{hughes2011aircraft}
Hughes, C., Van~Zante, D., and Heidmann, J., \enquote{Aircraft engine
  technology for green aviation to reduce fuel burn,} \emph{3rd AIAA
  Atmospheric Space Environments Conference}, 2011, p. 3531.

\bibitem[{Blakey et~al.(2011)Blakey, Rye, and Wilson}]{blakey2011aviation}
Blakey, S., Rye, L., and Wilson, C.~W., \enquote{Aviation gas turbine
  alternative fuels: A review,} \emph{Proceedings of the combustion institute},
  Vol.~33, No.~2, 2011, pp. 2863--2885.

\bibitem[{Gao and Mavris(2022)}]{gao2022statistics}
Gao, Z., and Mavris, D.~N., \enquote{Statistics and Machine Learning in
  Aviation Environmental Impact Analysis: A Survey of Recent Progress,}
  \emph{Aerospace}, Vol.~9, No.~12, 2022, p. 750.

\bibitem[{Pagoni and Psaraki-Kalouptsidi(2017)}]{pagoni2017calculation}
Pagoni, I., and Psaraki-Kalouptsidi, V., \enquote{Calculation of aircraft fuel
  consumption and CO2 emissions based on path profile estimation by clustering
  and registration,} \emph{Transportation Research Part D: Transport and
  Environment}, Vol.~54, 2017, pp. 172--190.

\bibitem[{Gao et~al.(2021)Gao, Behere, Li, Lim, Kirby, and
  Mavris}]{gao2021development}
Gao, Z., Behere, A., Li, Y., Lim, D., Kirby, M., and Mavris, D.~N.,
  \enquote{Development and Analysis of Improved Departure Modeling for Aviation
  Environmental Impact Assessment,} \emph{Journal of Aircraft}, Vol.~58, No.~4,
  2021, pp. 847--857.

\bibitem[{Ashok et~al.(2013)Ashok, Lee, Arunachalam, Waitz, Yim, and
  Barrett}]{ashok2013development}
Ashok, A., Lee, I.~H., Arunachalam, S., Waitz, I.~A., Yim, S.~H., and Barrett,
  S.~R., \enquote{Development of a response surface model of aviation's air
  quality impacts in the United States,} \emph{Atmospheric Environment},
  Vol.~77, 2013, pp. 445--452.

\bibitem[{Greenwood et~al.(2015)Greenwood, Schmitz, and
  Sickenberger}]{greenwood2015semiempirical}
Greenwood, E., Schmitz, F.~H., and Sickenberger, R.~D., \enquote{A
  semiempirical noise modeling method for helicopter maneuvering flight
  operations,} \emph{Journal of the American Helicopter Society}, Vol.~60,
  No.~2, 2015, pp. 1--13.

\bibitem[{Kang and Hansen(2018)}]{kang2018improving}
Kang, L., and Hansen, M., \enquote{Improving airline fuel efficiency via fuel
  burn prediction and uncertainty estimation,} \emph{Transportation Research
  Part C: Emerging Technologies}, Vol.~97, 2018, pp. 128--146.

\bibitem[{Baklacioglu(2016)}]{baklacioglu2016modeling}
Baklacioglu, T., \enquote{Modeling the fuel flow-rate of transport aircraft
  during flight phases using genetic algorithm-optimized neural networks,}
  \emph{Aerospace Science and Technology}, Vol.~49, 2016, pp. 52--62.

\bibitem[{Khan et~al.(2021)Khan, Ma, Ouyang, and Mo}]{khan2021prediction}
Khan, W.~A., Ma, H.-L., Ouyang, X., and Mo, D.~Y., \enquote{Prediction of
  aircraft trajectory and the associated fuel consumption using covariance
  bidirectional extreme learning machines,} \emph{Transportation Research Part
  E: Logistics and Transportation Review}, Vol. 145, 2021, p. 102189.

\bibitem[{Simone et~al.(2013)Simone, Stettler, and Barrett}]{simone2013rapid}
Simone, N.~W., Stettler, M.~E., and Barrett, S.~R., \enquote{Rapid estimation
  of global civil aviation emissions with uncertainty quantification,}
  \emph{Transportation Research Part D: Transport and Environment}, Vol.~25,
  2013, pp. 33--41.

\bibitem[{Allaire et~al.(2014)Allaire, Noel, Willcox, and
  Cointin}]{allaire2014uncertainty}
Allaire, D., Noel, G., Willcox, K., and Cointin, R., \enquote{Uncertainty
  quantification of an aviation environmental toolsuite,} \emph{Reliability
  Engineering \& System Safety}, Vol. 126, 2014, pp. 14--24.

\bibitem[{Gao et~al.(2019)Gao, Lim, Schwartz, and
  Mavris}]{gao2019nonparametric}
Gao, Z., Lim, D., Schwartz, K.~G., and Mavris, D.~N., \enquote{A
  nonparametric-based approach for the characterization and propagation of
  epistemic uncertainty due to small datasets,} \emph{AIAA Scitech 2019 forum},
  2019, p. 1490.

\bibitem[{Van~Pham et~al.(2010)Van~Pham, Tang, Alam, Lokan, and
  Abbass}]{van2010aviation}
Van~Pham, V., Tang, J., Alam, S., Lokan, C., and Abbass, H.~A.,
  \enquote{Aviation emission inventory development and analysis,}
  \emph{Environmental Modelling \& Software}, Vol.~25, No.~12, 2010, pp.
  1738--1753.

\bibitem[{Filippone and Parkes(2021)}]{filippone2021evaluation}
Filippone, A., and Parkes, B., \enquote{Evaluation of commuter airplane
  emissions: A European case study,} \emph{Transportation Research Part D:
  Transport and Environment}, Vol.~98, 2021, p. 102979.

\bibitem[{Huynh et~al.(2022)Huynh, Mahseredjian, and
  John~Hansman}]{huynh2022delayed}
Huynh, J.~L., Mahseredjian, A., and John~Hansman, R., \enquote{Delayed
  Deceleration Approach Noise Impact and Modeling Validation,} \emph{Journal of
  Aircraft}, 2022, pp. 1--13.

\bibitem[{Simons et~al.(2022)Simons, Besnea, Mohammadloo, Melkert, and
  Snellen}]{simons2022comparative}
Simons, D.~G., Besnea, I., Mohammadloo, T.~H., Melkert, J.~A., and Snellen, M.,
  \enquote{Comparative assessment of measured and modelled aircraft noise
  around Amsterdam Airport Schiphol,} \emph{Transportation Research Part D:
  Transport and Environment}, Vol. 105, 2022, p. 103216.

\bibitem[{Olive(2019)}]{olive2019traffic}
Olive, X., \enquote{Traffic, a toolbox for processing and analysing air traffic
  data,} \emph{Journal of Open Source Software}, Vol.~4, No.~39, 2019, pp.
  1518--1.

\bibitem[{Sun et~al.(2019)Sun, Ellerbroek, and Hoekstra}]{sun2019wrap}
Sun, J., Ellerbroek, J., and Hoekstra, J.~M., \enquote{WRAP: An open-source
  kinematic aircraft performance model,} \emph{Transportation Research Part C:
  Emerging Technologies}, Vol.~98, 2019, pp. 118--138.

\bibitem[{Kim et~al.(2022)Kim, Justin, Mavris, and Briceno}]{kim2022data}
Kim, J., Justin, C., Mavris, D., and Briceno, S., \enquote{Data-Driven Approach
  Using Machine Learning for Real-Time Flight Path Optimization,} \emph{Journal
  of Aerospace Information Systems}, Vol.~19, No.~1, 2022, pp. 3--21.

\bibitem[{Bellet et~al.(2013)Bellet, Habrard, and Sebban}]{bellet2013metric}
Bellet, A., Habrard, A., and Sebban, M., \enquote{A Survey on Metric Learning
  for Feature Vectors and Structured Data,} , 2013.
\newblock \doi{10.48550/ARXIV.1306.6709},
  \urlprefix\url{https://arxiv.org/abs/1306.6709}.

\bibitem[{Gao(2022)}]{gao2022representative}
Gao, Z., \enquote{Representative Data and Models for Complex Aerospace Systems
  Analysis,} Ph.D. thesis, Georgia Institute of Technology, 2022.

\bibitem[{Nuic et~al.(2010)Nuic, Poles, and Mouillet}]{nuic2010bada}
Nuic, A., Poles, D., and Mouillet, V., \enquote{BADA: An advanced aircraft
  performance model for present and future ATM systems,} \emph{International
  journal of adaptive control and signal processing}, Vol.~24, No.~10, 2010,
  pp. 850--866.

\bibitem[{ECAC-CEAC(December 2016)}]{Doc29}
ECAC-CEAC, \enquote{ECAC Doc 29 - Report on Standard Method of Computing Noise
  Contours around Civil Airports Volume 1: Applications Guide, 4th Edition,}
  Tech. rep., December December 2016.

\bibitem[{ECAC-CEAC'(July 2015)}]{Doc29-2}
ECAC-CEAC', \enquote{ECAC Doc 29 - Report on Standard Method of Computing Noise
  Contours around Civil Airports Volume 2: Technical Guide, 3rd Edition,} Tech.
  rep., July July 2015.

\bibitem[{Nuic(2010)}]{nuic2010user}
Nuic, A., \enquote{User manual for the Base of Aircraft Data (BADA) revision
  3.10,} \emph{Atmosphere}, Vol. 2010, 2010, p. 001.

\bibitem[{EUROCONTROL(March 2016)}]{bada4}
EUROCONTROL, \enquote{User Manual for the Base of Aircraft Data (BADA) Family
  4,} Technical Report EEC Technical/Scientific Report No. 12/11/22-58,
  EUROCONTROL Experimental Centre, March 2016.

\bibitem[{Baptista et~al.(2021)Baptista, Henriques, and
  Prendinger}]{baptista2021classification}
Baptista, M.~L., Henriques, E.~M., and Prendinger, H., \enquote{Classification
  prognostics approaches in aviation,} \emph{Measurement}, Vol. 182, 2021, p.
  109756.
\newblock \doi{10.1016/j.measurement.2021.109756}.

\bibitem[{Harrivel et~al.(2016)Harrivel, Liles, Stephens, Ellis, Prinzel, and
  Pope}]{harrivel2016psych}
Harrivel, A.~R., Liles, C., Stephens, C.~L., Ellis, K.~K., Prinzel, L.~J., and
  Pope, A.~T., \emph{Psychophysiological Sensing and State Classification for
  Attention Management in Commercial Aviation}, 2016.
\newblock \doi{10.2514/6.2016-1490}.

\bibitem[{D’Angelo and Rampone(2016)}]{dangelo2016classification}
D’Angelo, G., and Rampone, S., \enquote{Feature extraction and soft computing
  methods for aerospace structure defect classification,} \emph{Measurement},
  Vol.~85, 2016, pp. 192--209.
\newblock \doi{10.1016/j.measurement.2016.02.027}.

\bibitem[{Murca and Hansman(2018)}]{murca2018identification}
Murca, M. C.~R., and Hansman, R.~J., \enquote{Identification, characterization,
  and prediction of traffic flow patterns in multi-airport systems,} \emph{IEEE
  Transactions on Intelligent Transportation Systems}, Vol.~20, No.~5, 2018,
  pp. 1683--1696.

\bibitem[{Jensen et~al.(2017)Jensen, Thomas, Brooks, Brenner, and
  Hansman}]{jensen2017development}
Jensen, L., Thomas, J., Brooks, C., Brenner, M., and Hansman, R.~J.,
  \enquote{Development of Rapid Fleet-Wide Environmental Assessment
  Capability,} \emph{AIAA Modeling and Simulation Technologies Conference},
  2017, p. 3339.

\bibitem[{Gao et~al.(2022{\natexlab{a}})Gao, Puranik, and Mavris}]{gao2022prem}
Gao, Z., Puranik, T.~G., and Mavris, D.~N., \enquote{Probabilistic
  REpresentatives Mining (PREM): A Clustering Method for Distributional Data
  Reduction,} \emph{AIAA Journal}, Vol.~60, No.~4, 2022{\natexlab{a}}, pp.
  2580--2596.
\newblock \doi{10.2514/1.J061079}.

\bibitem[{Maruhashi et~al.(2022)Maruhashi, Grewe, Fr{\"o}mming, J{\"o}ckel, and
  Dedoussi}]{maruhashi2022transport}
Maruhashi, J., Grewe, V., Fr{\"o}mming, C., J{\"o}ckel, P., and Dedoussi,
  I.~C., \enquote{Transport Patterns of Global Aviation NOx and their
  Short-term O3 Radiative Forcing--A Machine Learning Approach,}
  \emph{Atmospheric Chemistry and Physics Discussions}, 2022, pp. 1--40.

\bibitem[{Li et~al.(2015)Li, Das, John~Hansman, Palacios, and
  Srivastava}]{li2015analysis}
Li, L., Das, S., John~Hansman, R., Palacios, R., and Srivastava, A.~N.,
  \enquote{Analysis of Flight Data Using Clustering Techniques for Detecting
  Abnormal Operations,} \emph{Journal of Aerospace Information Systems},
  Vol.~12, No.~9, 2015, pp. 587--598.
\newblock \doi{10.2514/1.I010329}.

\bibitem[{Puranik and Mavris(2018)}]{puranik2018anomaly}
Puranik, T.~G., and Mavris, D.~N., \enquote{Anomaly Detection in
  General-Aviation Operations Using Energy Metrics and Flight-Data Records,}
  \emph{Journal of Aerospace Information Systems}, Vol.~15, No.~1, 2018, pp.
  22--36.
\newblock \doi{10.2514/1.I010582}.

\bibitem[{Kumar et~al.(2021)Kumar, Corrado, Puranik, and
  Mavris}]{kumar2021classification}
Kumar, S.~G., Corrado, S.~J., Puranik, T.~G., and Mavris, D.~N.,
  \enquote{Classification and Analysis of Go-Arounds in Commercial Aviation
  Using ADS-B Data,} \emph{Aerospace}, Vol.~8, No.~10, 2021.
\newblock \doi{10.3390/aerospace8100291}.

\bibitem[{Tanguy et~al.(2016)Tanguy, Tulechki, Urieli, Hermann, and
  Raynal}]{tanguy2016natural}
Tanguy, L., Tulechki, N., Urieli, A., Hermann, E., and Raynal, C.,
  \enquote{Natural language processing for aviation safety reports: From
  classification to interactive analysis,} \emph{Computers in Industry},
  Vol.~78, 2016, pp. 80--95.
\newblock \doi{10.1016/j.compind.2015.09.005}, natural Language Processing and
  Text Analytics in Industry.

\bibitem[{Gao et~al.(2022{\natexlab{b}})Gao, Li, Puranik, and
  Mavris}]{gao2022minimax}
Gao, Z., Li, Y., Puranik, T.~G., and Mavris, D.~N., \enquote{Minimax and
  Multi-Criteria Selection of Representative Model Portfolios for Complex
  Systems Analysis,} \emph{AIAA Journal}, Vol.~60, No.~3, 2022{\natexlab{b}},
  pp. 1505--1521.
\newblock \doi{10.2514/1.J061007}.

\bibitem[{Aggarwal et~al.(2001)Aggarwal, Hinneburg, and
  Keim}]{aggarwal2001surprising}
Aggarwal, C.~C., Hinneburg, A., and Keim, D.~A., \enquote{On the Surprising
  Behavior of Distance Metrics in High Dimensional Spaces,} \emph{Proceedings
  of the 8th International Conference on Database Theory}, Springer-Verlag,
  Berlin, Heidelberg, 2001, p. 420–434.

\bibitem[{Corrado et~al.(2020)Corrado, Puranik, Pinon, and
  Mavris}]{corrado2020trajectory}
Corrado, S.~J., Puranik, T.~G., Pinon, O.~J., and Mavris, D.~N.,
  \enquote{Trajectory Clustering within the Terminal Airspace Utilizing a
  Weighted Distance Function,} \emph{Multidisciplinary Digital Publishing
  Institute Proceedings}, Vol.~59, 2020, p.~7.
\newblock \doi{10.3390/proceedings2020059007}.

\bibitem[{Gao et~al.(2022{\natexlab{c}})Gao, Kampezidou, Behere, Puranik,
  Rajaram, and Mavris}]{gao2022multi}
Gao, Z., Kampezidou, S.~I., Behere, A., Puranik, T.~G., Rajaram, D., and
  Mavris, D.~N., \enquote{Multi-level aircraft feature representation and
  selection for aviation environmental impact analysis,} \emph{Transportation
  Research Part C: Emerging Technologies}, Vol. 143, 2022{\natexlab{c}}, p.
  103824.

\bibitem[{Lee et~al.(2020)Lee, Thrasher, Hwang, Shumway, Zubrow, Hansen,
  Koopmann, and Solman}]{lee2020aviation}
Lee, C., Thrasher, T., Hwang, S., Shumway, M., Zubrow, A., Hansen, A.,
  Koopmann, J., and Solman, G., \enquote{Aviation Environmental Design Tool
  (AEDT) User Manual Version 3c,} Tech. rep., 2020.

\bibitem[{Johnson et~al.(1990)Johnson, Moore, and
  Ylvisaker}]{johnson1990minimax}
Johnson, M., Moore, L., and Ylvisaker, D., \enquote{Minimax and maximin
  distance designs,} \emph{Journal of Statistical Planning and Inference},
  Vol.~26, No.~2, 1990, pp. 131--148.
\newblock \doi{10.1016/0378-3758(90)90122-B}.

\bibitem[{Tan(2013)}]{tan2013minimax}
Tan, M. H.~Y., \enquote{Minimax Designs for Finite Design Regions,}
  \emph{Technometrics}, Vol.~55, No.~3, 2013, pp. 346--358.
\newblock \doi{10.1080/00401706.2013.804439}.

\bibitem[{Mak and Joseph(2018)}]{mak2018minimax}
Mak, S., and Joseph, V.~R., \enquote{Minimax and Minimax Projection Designs
  Using Clustering,} \emph{Journal of Computational and Graphical Statistics},
  Vol.~27, No.~1, 2018, pp. 166--178.
\newblock \doi{10.1080/10618600.2017.1302881}.

\bibitem[{Vanli et~al.(2012)Vanli, Zhang, Nguyen, and Wang}]{vanli2012minimax}
Vanli, O.~A., Zhang, C., Nguyen, A., and Wang, B., \enquote{A minimax sensor
  placement approach for damage detection in composite structures,}
  \emph{Journal of Intelligent Material Systems and Structures}, Vol.~23,
  No.~8, 2012, pp. 919--932.
\newblock \doi{10.1177/1045389X12440751}.

\bibitem[{Bien and Tibshirani(2011)}]{bien2011hierarchical}
Bien, J., and Tibshirani, R., \enquote{Hierarchical clustering with prototypes
  via minimax linkage,} \emph{Journal of the American Statistical Association},
  Vol. 106, No. 495, 2011, pp. 1075--1084.
\newblock \doi{10.1198/jasa.2011.tm10183}.

\bibitem[{Ao et~al.(2004)Ao, Yip, Ng, Cheung, Fong, Melhado, and
  Sham}]{ao2004clustag}
Ao, S.~I., Yip, K., Ng, M., Cheung, D., Fong, P.-Y., Melhado, I., and Sham,
  P.~C., \enquote{{CLUSTAG: hierarchical clustering and graph methods for
  selecting tag SNPs},} \emph{Bioinformatics}, Vol.~21, No.~8, 2004, pp.
  1735--1736.
\newblock \doi{10.1093/bioinformatics/bti201}.

\bibitem[{Kulis(2013)}]{kulis2013metric}
Kulis, B., \enquote{Metric learning: A survey,} \emph{Foundations and Trends in
  Machine Learning}, Vol.~5, No.~4, 2013, pp. 287--364.
\newblock \doi{10.1561/2200000019}.

\bibitem[{Mahalanobis(1936)}]{mahalanobis1936generalized}
Mahalanobis, P.~C., \enquote{On the generalized distance in statistics,}
  National Institute of Science of India, 1936.

\bibitem[{Ghojogh et~al.(2022)Ghojogh, Ghodsi, Karray, and
  Crowley}]{ghojogh2022metric}
Ghojogh, B., Ghodsi, A., Karray, F., and Crowley, M., \enquote{Spectral,
  Probabilistic, and Deep Metric Learning: Tutorial and Survey,} , 2022.
\newblock \doi{10.48550/ARXIV.2201.09267},
  \urlprefix\url{https://arxiv.org/abs/2201.09267}.

\bibitem[{Xing et~al.(2002)Xing, Jordan, Russell, and Ng}]{xing2002metric}
Xing, E., Jordan, M., Russell, S.~J., and Ng, A., \enquote{Distance Metric
  Learning with Application to Clustering with Side-Information,}
  \emph{Advances in Neural Information Processing Systems}, Vol.~15, edited by
  S.~Becker, S.~Thrun, and K.~Obermayer, MIT Press, 2002.

\bibitem[{Schultz and Joachims(2003)}]{Schultz2003distance}
Schultz, M., and Joachims, T., \enquote{Learning a Distance Metric from
  Relative Comparisons,} \emph{Advances in Neural Information Processing
  Systems}, Vol.~16, edited by S.~Thrun, L.~Saul, and B.~Sch\"{o}lkopf, MIT
  Press, 2003.

\bibitem[{Weinberger and Saul(2009)}]{weinberger2009distance}
Weinberger, K.~Q., and Saul, L.~K., \enquote{Distance metric learning for large
  margin nearest neighbor classification.} \emph{Journal of machine learning
  research}, Vol.~10, No.~2, 2009.

\bibitem[{Davis et~al.(2007)Davis, Kulis, Jain, Sra, and
  Dhillon}]{davis2007information}
Davis, J.~V., Kulis, B., Jain, P., Sra, S., and Dhillon, I.~S.,
  \enquote{Information-theoretic metric learning,} \emph{Proceedings of the
  24th international conference on Machine learning}, Association for Computing
  Machinery, 2007, pp. 209--216.
\newblock \doi{10.1145/1273496.1273523}.

\bibitem[{Weinberger and Saul(2008)}]{weinberger2008fast}
Weinberger, K.~Q., and Saul, L.~K., \enquote{Fast solvers and efficient
  implementations for distance metric learning,} \emph{Proceedings of the 25th
  international conference on Machine learning}, 2008, pp. 1160--1167.

\bibitem[{{de Vazelhes} et~al.(2020){de Vazelhes}, {Carey}, {Tang}, {Vauquier},
  and {Bellet}}]{metric-learn}
{de Vazelhes}, W., {Carey}, C., {Tang}, Y., {Vauquier}, N., and {Bellet}, A.,
  \enquote{metric-learn: {M}etric {L}earning {A}lgorithms in {P}ython,}
  \emph{Journal of Machine Learning Research}, Vol.~21, No. 138, 2020, pp.
  1--6.

\bibitem[{Van~der Maaten and Hinton(2008)}]{van2008visualizing}
Van~der Maaten, L., and Hinton, G., \enquote{Visualizing data using t-SNE,}
  \emph{Journal of machine learning research}, Vol.~9, No.~86, 2008, pp.
  2579--2605.

\bibitem[{Costa et~al.(2019)Costa, Ahamed, and Stephens}]{costa2019adaptive}
Costa, A.~C., Ahamed, T., and Stephens, G.~J., \enquote{Adaptive, locally
  linear models of complex dynamics,} \emph{Proceedings of the National Academy
  of Sciences}, Vol. 116, No.~5, 2019, pp. 1501--1510.
\newblock \doi{10.1073/pnas.1813476116}.

\end{thebibliography}

\end{document}